%% file: vil_arxiv.tex
\documentclass[10pt,twocolumn,letterpaper]{article}

\usepackage{iccv}
\usepackage{times}
\usepackage{epsfig}
\usepackage{graphicx}
\usepackage{amsmath}
\usepackage{amssymb}

\usepackage[pagebackref=true,breaklinks=true,letterpaper=true,colorlinks,bookmarks=false]{hyperref}

\usepackage{comment}
\usepackage{booktabs}       
\usepackage{multirow}

\usepackage{subfigure}
\usepackage{soul}

\usepackage{enumitem}
\usepackage[table]{xcolor}
\definecolor{Gray}{gray}{0.93}
\definecolor{Graylight}{gray}{0.95}
\definecolor{Grayheavy}{gray}{0.90}

\newcommand{\vil}{\text{ViL}}

\usepackage{pifont}
\newcommand{\cmark}{\text{\ding{51}}}
\newcommand{\xmark}{\text{\ding{55}}}

\newcommand\blfootnote[1]{%
  \begingroup
  \renewcommand\thefootnote{}\footnote{#1}%
  \addtocounter{footnote}{-1}%
  \endgroup
}

\iccvfinalcopy 

\def\iccvPaperID{11316} 

\ificcvfinal\fi

\begin{document}

\title{Multi-Scale Vision Longformer: \\
A New Vision Transformer for High-Resolution Image Encoding}


\author{
Pengchuan Zhang\textsuperscript{$\heartsuit$} 
\and Xiyang Dai\textsuperscript{$\heartsuit\dagger$}
\and Jianwei Yang\textsuperscript{$\heartsuit\dagger$}
\and Bin Xiao\textsuperscript{$\heartsuit$} 
\and Lu Yuan\textsuperscript{$\heartsuit$} 
\and Lei Zhang\textsuperscript{$\heartsuit$} 
\and Jianfeng Gao\textsuperscript{$\heartsuit$}
}

\maketitle

\blfootnote{\textsuperscript{$\heartsuit$}Microsoft Corporation
\hspace{10mm}
 $\dagger$ indicates equal contributions.}

\begin{abstract}
  This paper presents a new Vision Transformer (ViT) architecture \emph{Multi-Scale Vision Longformer}, which significantly enhances the ViT of ~\cite{dosovitskiy2020image} for encoding high-resolution images using two techniques.
   The first is the multi-scale model structure, which provides image encodings at multiple scales with manageable computational cost.
   The second is the attention mechanism of Vision Longformer, which is a variant of Longformer~\cite{beltagy2020longformer}, originally developed for natural language processing, and achieves a linear complexity w.r.t. the number of input tokens. 
   A comprehensive empirical study shows that the new ViT significantly outperforms several strong baselines, including the existing ViT models and their ResNet counterparts, and the Pyramid Vision Transformer from a concurrent work~\cite{wang2021pyramid}, on a range of vision tasks, including image classification, object detection, and segmentation. The models and source code are released at \url{https://github.com/microsoft/vision-longformer}.

\end{abstract}

\input{intro}

\input{relatedwork}

\input{msvit}

\input{detection}

\section{Experiments}

\input{classification_exps}

\input{detection_exps}

\input{ablation_studies}

\input{conclusion}

{\small
\bibliographystyle{ieee_fullname}
\bibliography{egbib}
}

\appendix
\clearpage
\newpage
\input{app_settings}
\input{appendix}
\input{app_vil_implement}
\input{app_random_shift}
\input{app_otherattns}

\end{document}

%% file: intro.tex
\section{Introduction}
Vision Transformer (ViT)~\cite{dosovitskiy2020image} has shown promising results on image classification tasks for its strong capability of long range context modeling. But its quadratic increase of both computational and memory complexity hinders its application on many vision tasks that require high-resolution feature maps computed on high-resolution images\footnote{In this paper, encoding a high-resolution image means generating high-resolution feature maps for high-resolution images. 
}, 
like object detection~\cite{ren2015faster, lin2017focal}, segmentation~\cite{long2015fully, chen2017deeplab}, and human pose estimation~\cite{xiao2018simple, sun2019deep}. Vision-language tasks, like VQA, image captioning, and image-text retrieval, also benefit from high-resolution feature maps~\cite{jiang2020defense,zhang2021vinvl}, which are extracted with pre-trained CNN models. 
Developing a vision Transformer that can process high-resolution feature maps is a critical step toward the goal of unifying the model architecture of vision and language modalities and improving  multi-modal representation learning.

In this paper, we propose a new vision Transformer architecture \emph{Multi-Scale Vision Longformer}, which significantly enhances the baseline ViT~\cite{dosovitskiy2020image} for encoding high-resolution images using two techniques: (1) the multi-scale model structure, and (2) the attention mechanism of Vision Longformer. 

Models with multi-scale (pyramid, hierarchical) structure provide a comprehensive encoding of an image at multiple scales, while keeping the computation and memory complexity manageable. 
Deep convolutional networks are born with such multi-scale structure, which however is not true for the conventional ViT architecture. 
To obtain a multi-scale vision Transformer, we stack multiple (e.g., four) vision Transformers (ViT stages) sequentially. 
The first ViT stage operates on a high-resolution feature map but has a small hidden dimension. As we go to later ViT stages, the feature map resolution reduces while the hidden dimension increases. The resolution reduction is achieved by performing patching embedding at each ViT stage. 
In our experiments, we find that with the same number of model parameters and the same model FLOPs, the multi-scale ViT achieves a significantly better accuracy than the vanilla ViT on image classification task. 
The results show that the multi-scale structure not only improves the computation and memory efficiency, but also boosts the classification performance.
The proposed multi-scale ViT has the same network structure as conventional (multi-scale) CNN models such as ResNet~\cite{he2016deep}, and can serve as a replace-and-plug-in choice for almost all ResNet applications. In this paper, we demonstrate this plausible property in image classification, object detection and instance segmentation.

The multi-scale structure alone is not sufficient to scale up ViT to process high-resolution images and feature maps, due to the quadratic increase of the computation and memory complexity with respect to the number of tokens in the self-attention layers. Compared to natural language tasks where data is 1-D, this problem is more severe in vision tasks where the increase in complexity is quartic (fourth order) with the increase of image resolution. 
For example, the computational complexity of a $4\times$ higher resolution multi-head self attention (MSA) layer (hidden dimension reduced by 4, i.e., $4H\times 4W\times \frac{D}{4}$) equals to that of 64 layers in the original size (i.e., $H\times W\times D$). To address this challenge, we develop a 2-D version of Longformer\cite{beltagy2020longformer}, called \emph{Vision Longformer}, to achieve a linear complexity w.r.t. the number of tokens (quadratic w.r.t. resolution). 
Our experiments show that compared to the baseline ViT, Vision Longformer shows no performance drop while significantly reduces the computational and memory cost in encoding images. 
The result indicates that the ``local attention + global memory" structure in Vision Longformer is a desirable inductive bias for vision Transformers. 
We also compare Vision Longformer with other efficient attention mechanisms. The result again validates its superior performance on both image classification and object detection tasks. 

The main contributions of this paper are two-fold: (1) We propose a new vision Transformer that uses the multi-scale model structure and the attention mechanism of 2-D Longformer for efficient high-resolution image encoding.
(2) We perform a comprehensive empirical study to show that the proposed ViT significantly outperforms strong baselines, including previous ViT models, their ResNet counterparts, and a model from a concurrent work, on image classification, object detection and segmentation tasks.

%% file: relatedwork.tex
\section{Related Work}
The Vision Transformer (ViT)~\cite{dosovitskiy2020image} applies a standard Transformer, originally developed for natural language processing (NLP), for image encoding by treating an image as a word sequence, i.e., splitting an image into patches (words) and using the linear embeddings of these patches as an input sequence.
ViT has shown to 
outperform convolution neural network (CNN) models such as the ResNet~\cite{he2016deep}, achieving state-of-the-art performance on multiple image classification benchmarks, where training data is sufficient.
DeiT~\cite{touvron2020training} is another computer vision model that leverages Transformer. It uses a teacher-student strategy specific to Transformers to improve data efficiency in training. Thus, compared to ViT, it requires much less training data and computing resources to produce state-of-the-art image classification results. 
In addition to image classification, Transformers have also been  applied to other compute vision tasks, including object detection~\cite{carion2020end,zhu2020deformable,zheng2020end,dai2020up}, segmentation~\cite{wang2020max,wang2020end}, 
image enhancement~\cite{chen2020pre,yang2020learning}, image generation~\cite{parmar2018image,chen2020generative}, video processing~\cite{zeng2020learning,Zhou_2018_CVPR}, and vision-language tasks~\cite{lu2019vilbert,tan2019lxmert,chen2019uniter,su2019vl,li2019visualbert,li2019unicoder,zhou2019unified,li2020oscar}.

Developing an efficient attention mechanism for high-resolution image encoding is the focus of this work. 
Our model is inspired by the efficient attention mechanisms developed for Transformers, most of which are for NLP tasks. 
These mechanisms can be grouped into four categories.
The first 
is the sparse attention mechanism, including content-independent sparsity~\cite{parmar2018image,child2019generating,qiu2019blockwise,ho2019axial} and content-dependent sparsity~\cite{kitaev2020reformer,roy*2020efficient,tay2020sparse,zhou2020informer}.
Axial Transformer~\cite{ho2019axial} and Image Transformer~\cite{parmar2018image} are among few sparsity-based efficient attentions that are developed for image generation.
The second is the memory-based mechanism, including Compressive Transformers~\cite{rae2019compressive} and Set Transformer~\cite{lee2019set}. 
These models use some extra global tokens as static memory and allow all the other tokens to attend only to those global tokens. 
The third is the low-rank based mechanism. For example the Linformer~\cite{wang2020linformer}
projects the input key-value pairs into a smaller chunk, and performs cross-attention between the queries and the projected key-value pairs. 
The fourth is the (generalized) kernel-based mechanism, including Performer\cite{choromanski2020rethinking} and Linear Transformers\cite{katharopoulos2020transformers}. 
Many models utilize hybrid attention mechanisms. 
For example, Longformer\cite{beltagy2020longformer}, BigBird\cite{zaheer2020big} and ETC\cite{ainslie2020etc} combine the sparsity and memory mechanisms; Synthesizers\cite{tay2020synthesizer} combines the sparsity and low-rank mechanisms. 
Readers may refer to \cite{tay2020efficient} and \cite{tay2020long} for a comprehensive survey and benchmarks, respectively. 

In this paper, we developed a 2-D version of Longformer\cite{beltagy2020longformer}, called Vision Longformer, which utilizes both the sparsity and memory mechanisms. Its conv-like sparsity mechanism is conceptually similar to 
the sparsity mechanism used in the Image Transformer\cite{parmar2018image}.
The multi-scale vision Transformer architecture is another technique we use in our proposed high-resolution Vision Longformer. 
The hierarchical Transformers~\cite{Pappagari2019} for NLP contain two stages, with the first stage processing overlapping segments and the second stage using the embeddings of the \texttt{CLS} tokens from all segments as input. 
In our proposed Vision Longformer, size reduction is performed by the patch embedding at the beginning of each stage, by merging all tokens in a patch from previous stage into a single token at the current stage. We typically use 4 stages for our model since we have empirically verified that using 4 stages is better than using 2 or 3 stages, especially for object detection tasks. Informer\cite{zhou2020informer} takes a similar stacked multi-stage approach to encoding long sequences, where the size reduction between stages is achieved by max-pooling. 

Pyramid Vision Transformer (PVT)~\cite{wang2021pyramid}, Swin Transformer~\cite{liu2021swin} and HanoNet~\cite{vaswani2021scaling} are concurrent works of ours. 
All these works use a multi-scale architecture where multiple (slightly modified) ViTs are stacked. 
The authors of PVT propose the spatial-reduction
attention (SRA) to alleviate the cost increase in self-attention layers. However, the computation and memory complexity of PVT still increases quartically w.r.t. resolution (with a much smaller constant). 
Swin Transformer~\cite{liu2021swin} and HanoNet~\cite{vaswani2021scaling} utilizes similar local attention mechanism as our Vision Longformer, but from different perspectives and implementations.

%% file: msvit.tex
\section{Multi-Scale Stacked Vision Transformers}
\begin{figure}[t!]
\includegraphics[width=0.48\textwidth]{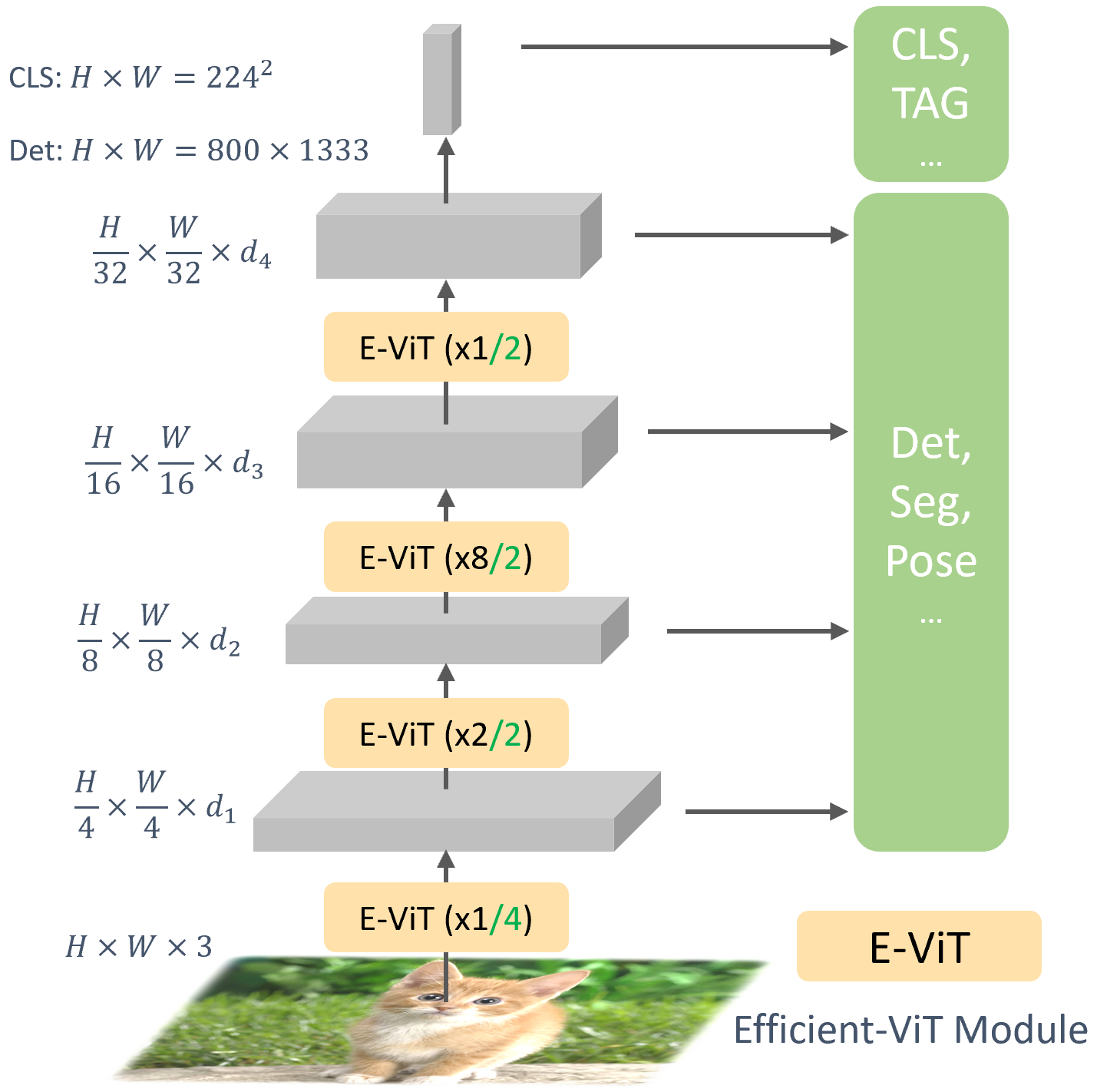}\\
\includegraphics[width=0.48\textwidth]{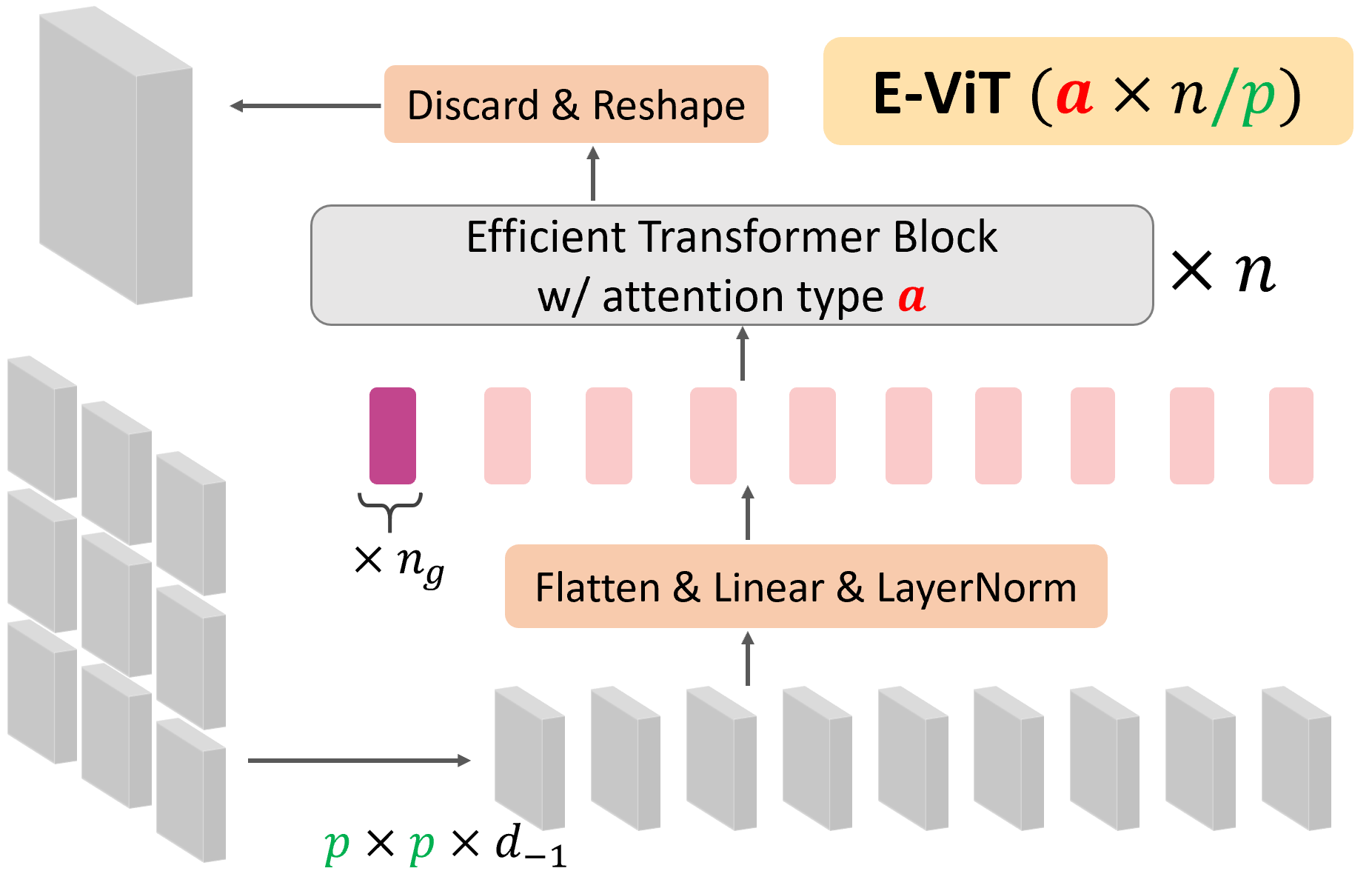}
\caption{A Multi-scale vision Transformers (bottom) by stacking 4 E-ViT modules (Top). An E-ViT ($\textcolor{red}{a}\times n \textcolor{green}{~/~p}$) module is a ViT encoder with an efficient attention mechanism $\textcolor{red}{a}$, $n$ efficient transformer blocks, input patch size $\textcolor{green}{p}$. We add a LayerNorm after the patch embedding. We add $n_g$ extra global tokens, as a form of global memory, and simply throw them away when going to the next stage.}
\label{fig:ms_vit}
\vspace{-2mm}
\end{figure}

\subsection{Multi-Scale Model Architecture}
\label{subsec:msvit}

\paragraph{Efficient ViT (E-ViT).}
As shown in Figure~\ref{fig:ms_vit} (Bottom), 
we improve the encoding efficiency of vision Transformer by making the following modifications to the vanilla ViT. The modified ViT is referred to as Efficient ViT (E-ViT).
\begin{enumerate}
    \item We add a Layer Normalization (LayerNorm) after the patch embedding.
    \item We define a number of \emph{global tokens}, including the \texttt{CLS} token. Correspondingly, the tokens associated with image and feature patches are referred to as \emph{local tokens} afterwards. 
    \item We replace the vanilla full self-attention with an efficient attention mechanism, denoted by $\textcolor{red}{a}$, which will be described in detail in Sections~\ref{subsec:2dlongformer} and \ref{subsec:efficientattn}.
    \item We use either an Absolute 2-D Positional Embedding (APE for short, separately encoding $x$ and $y$ coordinates and concatenating them) or a Relative Positional Bias (RPB for short) to replace the original absolute 1-D positional embedding.
\end{enumerate}
Except for attention $\textcolor{red}{a}$, E-ViT has the following architecture parameters inherited from the vanilla ViT~: input patch size $\textcolor{green}{p}$, number of attention blocks $n$, hidden dimension $d$ and number of heads $h$, denoted as $\text{E-ViT}(a\times n/p~;h,d,n_g)$. 
Using the full attention mechanism (i.e., \textcolor{red}{$a$} = full) and one global token (i.e., the \texttt{CLS} token with $n_g = 1$), the deficient $\text{E-ViT} (\text{full}\times 12/16~;h,d,1)$ models still achieve better ImageNet classification performance than the baseline ViT for both tiny ($h=3, d=192$) and small ($h=6, d=384$) model sizes, as shown in Table~\ref{tab:flatvsmultiscale}. 
The performance gain is attributed to the added LayerNorm, as we show in the Supplementary. 

Mathematically, an $\text{E-ViT}(a\times n/p~;h,d,n_g)$ encoding module can be written as:
\begin{align}
    z_0 &= [x_{g}^1;\dots;x_{g}^{n_g}; LN(x_p^1 E); \dots; LN(x_p^{n_l} E)] + E_{ops}, \label{eq:patchembed}\\
    z'_k &= MSA_{\textcolor{red}{a}}(LN(z_{k-1})) + z_{k-1}, \quad k=1,..,n \label{eq:emsa}\\
    z_k &= MLP(LN(z'_{k})) + z'_{k}, \quad k=1,..,n, \label{eq:ffn}
\end{align}
where $LN$ is the added Layer Normalization after the patch embedding $E$, $MSA_{\textcolor{red}{a}}$ is the multi-head self-attention with attention type $\textcolor{red}{a}$, and $MLP$ is the feed-forward block in a standard Transformer. When the absolute 2-D positional embedding is used, $E_{ops} \in \mathbb{R}^{(n_l + n_g)\times d}$ contains the 2-D positional embedding of $n_l$ local tokens and the 1-D positional embedding of $n_g$ global tokens. When the relative positional bias is used, $E_{ops} = 0$ and the per-head relative positional bias is directly added to the attention scores in the $MSA_{\textcolor{red}{a}}$ modules, as in Equation~\eqref{eq:rpb}. 

\noindent\textbf{Stack multiple E-ViT modules as multi-scale vision Transformers.}
As illustrated in Figure~\ref{fig:ms_vit} (Top), 
a multi-scale Vision Transformer is built by stacking multiple E-ViT modules (or stages). 
In what follows, we describe several design choices we have made when building the multi-scale ViT. 

\noindent\textbf{What are the patch size and hidden dimension at each stage?} 
As required in object detection and human pose estimation, for models with 4-scale feature maps, the first feature map needs to down-sample the image by 4 and thus stage 1 can be written as $\text{E-ViT}(a_1\times n_1/4~;h_1,d_1,n_{g,1})$. 
We typically use only one attention block, i.e., $n_1 = 1$. The first stage generates the highest-resolution feature map, which consumes lots of memory, as shown in Table~\ref{tab:flatvsmultiscale}. 
We also construct several 3-stage models, whose first stage patch size is 8. 
For later stages, the patch sizes are set to 2, which downsizes the feature map resolution by 2. Following the practice in ResNet, we increase the hidden dimension twice when downsizing the feature map resolution by 2.
We list a few representative model configurations in Table~\ref{tab:model_arch}. 
Different attention types ($\textcolor{red}{a}$) have different choices of number of global tokens $n_g$. But they share the same model  configurations. Thus we do not specify $\textcolor{red}{a}$ and $n_g$ in Table~\ref{tab:model_arch}. 
Please refer to the Supplementary for the complete list of model configurations used in this paper,

\begin{table}[ht]
\begin{center}
\resizebox{\linewidth}{!}{
\begin{tabular}{l@{\hspace{1.5pt}}|c@{\hspace{3pt}}|c@{\hspace{3pt}}|c@{\hspace{3pt}}|c@{\hspace{3pt}}}
\toprule
\multirow{2}{*}{Size} & Stage1 & Stage2 & Stage3 & Stage4 \\
 & n,p,h,d & n,p,h,d & n,p,h,d & n,p,h,d \\
\midrule
Tiny & 1,4,1,48 & 1,2,3,96 & 9,2,3,192 & 1,2,6,384 \\
\hline
Small & 1,4,3,96 & 2,2,3,192 & 8,2,6,384 & 1,2,12,768 \\
\hline
Medium & 1,4,3,96 & 4,2,3,192 & 16,2,6,384 & 1,2,12,768 \\
\hline
Base & 1,4,3,96 & 8,2,3,192 & 24,2,6,384 & 1,2,12,768 \\
\midrule
\hline
Small-3stage & \multicolumn{2}{c|}{2,8,3,192} & 9,2,6,384 & 1,2,12,768 \\
\bottomrule
\end{tabular}
}
\end{center}
\caption{Model architecture for multi-scale stacked ViTs. Architecture parameters for each E-ViT stage$\text{E-ViT}(a\times n/p~;h,d)$: number of attention blocks $n$, input patch size $p$, number of heads $h$ and hidden dimension $d$. See the meaning of these parameters in Figure~\ref{fig:ms_vit} (Bottom).}
\label{tab:model_arch}
\vspace{-2mm}
\end{table}

\noindent\textbf{How to connect global tokens between consecutive stages?} The choice varies at different stages and among different tasks. 
For the tasks in this paper, e.g., classification, object detection, instance segmentation, we simply discard the global tokens and only reshape the local tokens as the input for next stage. In this choice, global tokens only plays a role of an efficient way to globally communicate between distant local tokens, or can be viewed as a form of global memory. These global tokens are useful in vision-language tasks, in which the text tokens serve as the global tokens and will be shared across stages. 

\noindent\textbf{
Should we use the average-pooled layer-normed features or the LayerNormed \texttt{CLS} token's feature for image classification?} 
The choice makes no difference for flat models. 
But the average-pooled feature performs better than the \texttt{CLS} feature for multi-scale models, especially for the multi-scale models with only one attention block in the last stage (including all models in Table~\ref{tab:model_arch}). 
Please refer to the Supplementary for an ablation study.

As reported in Table~\ref{tab:flatvsmultiscale}, the multi-scale models outperform the flat models even in low-resolution classification tasks, demonstrating the importance of multi-scale structure. However, the full self-attention mechanism suffers from the quartic computation/memory complexity w.r.t. the resolution of feature maps, as shown in Table~\ref{tab:flatvsmultiscale}.
Thus, it is impossible to train 4-stage multi-scale ViTs with full attention using the same setting (batch size and hardware) used for DeiT training.

\begin{table}[ht]
\begin{center}
\resizebox{\linewidth}{!}{
\begin{tabular}{l@{\hspace{3pt}}|c@{\hspace{3pt}}|c|c|c}
\toprule
\multirow{2}{*}{Model} & \#Params & FLOPs & Memory & Top-1 \\
 & (M) & (G) & (M) & (\%) \\
\midrule
DeiT-Small~/~16~\cite{touvron2020training} & 22.1 & 4.6 & 67.1 & 79.9 \\
$\text{E-ViT}(\text{full}/16)$-APE & 22.1 & 4.6 & 67.1 & 80.4/80.7 \\
\hline
Full-1,10,1-APE & 27.58 & 4.84 & 78.5 & 81.7 \\
Full-2,9,1-APE & 26.25 & 5.05 & 93.8 & 81.7 \\
Full-1,1,9,1-APE & 25.96 & 6.74 & 472.9 & 81.9  \\
Full-1,2,8,1-APE & 24.63 & 6.95 & 488.3 & 81.9 \\
\hline
\vil-1,10,1-APE & 27.58 & 4.67 & 73.0 & 81.6 \\
\vil-2,9,1-APE & 26.25 & 4.71 & 81.4 & 81.8 \\
\vil-1,1,9,1-APE & 25.96 & 4.82 & 108.5 & 81.8  \\
\vil-1,2,8,1-APE & 24.63 & 4.86 & 116.8 & 82.0 \\
\hline
\vil-1,10,1-RPB & 27.61 & 4.67 & 78.8 & 81.9 \\
\vil-2,9,1-RPB & 26.28 & 4.71 & 88.7 & 82.3 \\
\vil-1,1,9,1-RPB & 25.98 & 4.82 & 121.8 & 82.2  \\
\vil-1,2,8,1-RPB & 24.65 & 4.86 & 131.6 & 82.4 \\
\bottomrule
\end{tabular}
}
\end{center}
\caption{Flat vs Multi-scale Models: Number of paramers, FLOPS, memory per image (with Pytorch Automatic Mixed Precision enabled), and ImageNet accuracy with image size 224. ``Full-2,9,1-APE'' stands for a 3-stage multiscale ViT with $a=\text{full}$ attention, with 2,9,1 number of attention blocks in each stage, respectively, and with Absolute 2-D Positional Embedding (APE). Since all our multi-scale models use average-pooled feature from the last stage for classification, we report Top-1 accuracy of ``$\text{E-ViT}(\text{full}/16)$-APE'' both with the CLS feature (first) and with the average-pooled feature (second). The multi-scale models consistently outperform the flat models, but the memory usage of full attention quickly blows up when only one high-resolution block is introduced. The Vision Longformer (``\vil-") saves FLOPs and memory, without performance drop. Using relative positional bias (``\vil-***-RPB") further improves the performance.}
\label{tab:flatvsmultiscale}
\vspace{-3mm}
\end{table}

\begin{figure}[t!]
\includegraphics[width=0.45\textwidth]{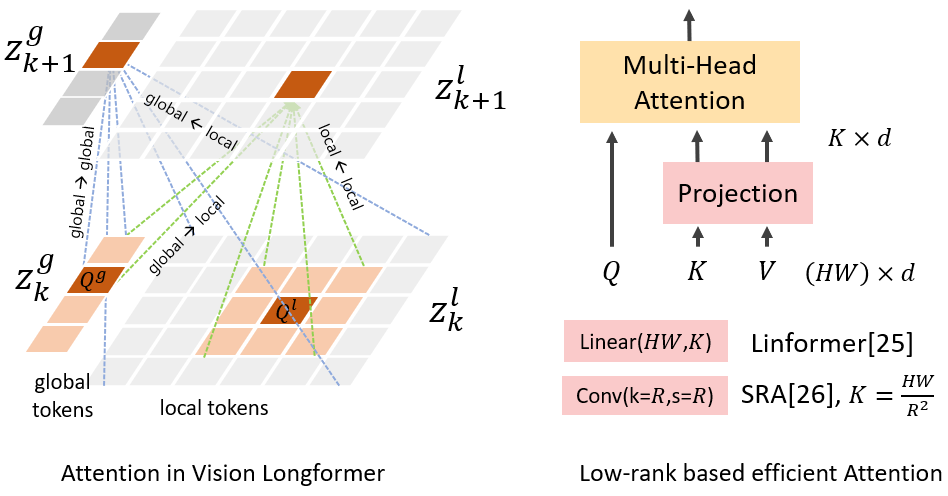}
\caption{Left: the Vision Longformer attention mechanism. Right: the Low-rank based attention mechanism. Without ``local$\to$local" attentions in Vision Longformer, we get the Global Former. With a linear layer as the projection, we get Linformer\cite{wang2020linformer}. With a conv layer with equal kernel size and stride, we get Spatial Reduction Attention (SRA)\cite{wang2021pyramid}.}
\label{fig:vil_block}
\vspace{-2mm}
\end{figure}

\subsection{Vision Longformer: A ``Local Attention + Global Memory" Mechanism}
\label{subsec:2dlongformer}
We propose to use the "local attention + global memory" efficient mechanism, as illustrated in Figure~\ref{fig:vil_block} (Left), to reduce the  computational and memory cost in the E-ViT module. 
The 2-D Vision Longformer is an extension of the 1-D Longformer~\cite{beltagy2020longformer} originally developed for NLP tasks. 
We add $n_g$ global tokens (including the \texttt{CLS} token) that are allowed to attend to all tokens, serving as global memory. 
Local tokens are allowed to attend to only global tokens and their local 2-D neighbors within a window size. 
After all, there are four components in this "local attention + global memory" mechanism,
namely global-to-global, local-to-global, global-to-local, and local-to-local, 
as illustrated in Figure~\ref{fig:vil_block} (Left). 
In Equation~\eqref{eq:emsa}, a Multi-head Self-Attention (MSA) block with the Vision Longformer attention mechanism is denoted as $MSA_{\textcolor{red}{\vil}}$, i.e., $a = \vil$ in Equation~\eqref{eq:emsa}.

\noindent\textbf{Relative positional bias for Vision Longformer.} Following \cite{raffel2019exploring,bao2020unilmv2,liu2021swin}, we add a relative positional bias $B$ to each head when computing the attention score:
\begin{equation}\label{eq:rpb}
    \text{Attention}(Q,K,V) = \text{SoftMax}(Q K^T/\sqrt{d} + B) V,
\end{equation}
where $Q,K,V$ are the query, key and value matrices and $d$ is the query/key dimension. This relative positional bias makes Vision Longformer translational invariant, which is a desired property for vision models. We observe significant improvements over the absolute 2-D positional embedding, as shown in Table~\ref{tab:flatvsmultiscale} for ImageNet classification and Section~\ref{subsec:ablation_classification} for COCO object detection.

\noindent\textbf{Theoretical complexity.} Given the numbers of global and local tokens, denoted by $n_g$ and $n_l$ respectively, and local attention window size $w$, the memory complexity of the $MSA_{\textcolor{red}{\vil}}$ block is $\mathcal{O}(n_g (n_g + n_l) + n_l w^2)$. Although \cite{beltagy2020longformer} points out that separating the attention parameters for global and local tokens is useful, we do not observe obvious gain in our experiments and thus simply let them share the same set of attention parameters. 
We empirically set the window size $w$ to 15 for all E-ViT stages, which makes our model comparable with the global attention window size 14 of ViT/16 acted on $224\times 224$ images. 
With such a window size, only attentions in the first two stages (in 4-stage multi-scale ViTs) are local. 
The attentions in the later two stages are equivalent\footnote{Equivalent in our sliding chunks implementation, which is our default choice.} to full attention.
In our experiments,
we find that it is sufficient to use only one global token ($n_g=1$) for ImageNet classification problems. 
So, the effective memory complexity of the $MSA_{\textcolor{red}{\vil}}$ block is $\mathcal{O}((15^2 + 1)n_l)$, which is linear w.r.t. the number of tokens.

\paragraph{Superior performance in ImageNet classification.} 
Results in Table~\ref{tab:flatvsmultiscale} show that in comparison with the full attention models, the proposed multi-scale Vision Longformer achieves a similar or even better performance, while saving significant memory and computation cost. The memory saving is significant for feature maps with resolution $56 \times 56$ (i.e., the feature maps in the first stage of a 4-stage multi-scale model). The savings are even more significant for higher resolution feature maps.
This makes Vision Longformer scalable to high-resolution vision tasks, such as object detection and segmentation. When equipped with relative positional bias, Vision Longformer outperforms the full attention models with absolute positional embedding. This indicates that the ``local attention + global memory" mechanism is a good inductive bias for vision Transformers. 

\noindent\textbf{Three implementations of Vision Longformer and its random-shifting training strategy.}
Vision Longformer is conceptually similar to conv-like local attention. 
We have implemented Vision Longformer in three ways: (1) using Pytorch's unfold function (nn.unfold or tensor.unfold), (2) using a customized CUDA kernel and (3) using a sliding chunk approach. 
The unfold implementation is simple but very slow, i.e., 24 times slower than full attention on $40 \times 40 \times 768$ feature map. 
The implementation using the customized CUDA kernel is about 20\% faster than the full attention in the same setting, while achieving the theoretical memory complexity. The sliding-chunk approach is the fastest, which is 60\% faster than the full attention with a cost of consuming slightly more memory than the theoretical complexity. With the sliding chunk implementation, we also propose a random-shifting training strategy for Vision Longformer, which further improves the training speed and memory consumption during training.  
Please refer to the Supplementary for details of these implementations and the random-shifting training strategy.

\subsection{Other Efficient Attention Mechanisms}
\label{subsec:efficientattn}
We compare Vision Longformer with the following alternative choices of efficient attention methods. We put detailed descriptions of these methods and their experimental setup in the Supplementary. 

\noindent\textbf{Pure global memory ($\textcolor{red}{a} = \text{global}$).} In Vision Longformer, see Figure~\ref{fig:vil_block} (Left), if we remove the local-to-local attention, then we obtain the pure global memory attention mechanism (called Global Attention hereafter). Its memory complexity is $\mathcal{O}(n_g(n_g + n_l))$, which is also linear w.r.t. $n_l$. However, for this pure global memory attention, $n_g$ has to be much larger than 1. We gradually increase $n_g$ (by 2 each time) and its performance gets nearly saturated at 128. Therefore, $n_g = 128$ is the default for this Global attention. 

\noindent\textbf{Linformer\cite{wang2020linformer} ($\textcolor{red}{a} = \text{LIN}$)} projects the $n_l \times d$ dimensional keys and values to $K \times d$ dimensions using additional projection layers, where $K \ll n_l$. Then the $n_l$ queries only attend to these projected $K$ key-value pairs. The memory complexity of Linformer is $\mathcal{O}(K n_l)$. We gradually increase $K$ (by 2 each time) and its performance gets nearly saturated at 256. Therefore, $K = 256$ is the default for this Linformer attention, which turns out to be the same with the recommended value. Notice that Linformer's projection layer (of dimension $K \times n_l$) is specific to the current $n_l$, and cannot be transferred to higher-resolution tasks that have a different $n_l$.

\noindent\textbf{Spatial Reduction Attention (SRA)~\cite{wang2021pyramid} ($\textcolor{red}{a} = \text{SRA}$)} is similar to Linformer, but uses a convolution layer with kernel size $R$ and stride $R$ to project the key-value pairs, hence resulting in $n_l / R^2$ compressed key-value pairs. Therefore, The memory complexity of SRA is $\mathcal{O}(n_l^2/R^2)$, which is still quadratic w.r.t. $n_l$ but with a much smaller constant $1/R^2$. When transferring the ImageNet-pretrained SRA-models to high-resolution tasks, SRA still suffers from the quartic computation/memory blow-up w.r.t. the feature map resolution. Pyramid Vision Transformer~\cite{wang2021pyramid} uses this SRA to build multi-scale vision transformer backbones, with different spatial reduction ratios ($R_1=8, R_2=4,R_3=2, R_4=1$) for each stage. With this PVT's setting, the key and value feature maps at all stages are essentially with resolution $H/32 \times W/32$. 

\noindent\textbf{Performer~\cite{choromanski2020rethinking} ($\textcolor{red}{a} = \text{performer}$)} uses random kernels to approximate the Softmax computation in MSA, and achieves a linear computation/memory complexity with respect to $n_l$ and the number of random features. We use the default 256 orthogonal random features (OR) for Performer, and provide other details in the Supplementary. 

\begin{table}[!ht]
\begin{center}
\resizebox{\linewidth}{!}{
\footnotesize
\begin{tabular}{l@{\hspace{3pt}}|c@{\hspace{5pt}}c|c@{\hspace{6pt}}c|c@{\hspace{3pt}}}
\toprule
Multi-scale & \multicolumn{2}{c|}{Tiny-4stage~/~4} & \multicolumn{2}{c|}{Small-4stage~/~4} & Trans\\ 
Models & 1,1,9,1 & 1,2,8,1 & 1,1,9,1 & 1,2,8,1 & 2Det  \\ 
\midrule
Full & $76.06$ & 75.60 & 81.93 & 81.91 & -- \\
\vil & 76.18 & 75.98 & 81.79 & 81.99 & \cmark \\
\midrule
Global & 71.52 & 72.00  & 79.17 & 78.97 & \cmark \\
Linformer~\cite{wang2020linformer} & 74.71 & 74.74 & 81.19 & 80.98 & \xmark \\
SRA/64\cite{wang2021pyramid}  & 69.08 & 68.78 & 76.35 & 76.37 & \cmark \\
SRA/32\cite{wang2021pyramid}  & 73.22 & 73.2 & 79.96 & 79.9 & -- \\
Performer & 71.12 & 73.09 & 78.81 & 78.72 & \cmark \\
\midrule
Par-Global & 75.32 & 75.4 & 81.6 & 81.46 & -- \\
Par-Linformer & 75.56 & 75.33 & 81.66 & 81.79 & \xmark \\
Par-SRA/32  & 75.2 & 75.26 & 81.62 & 81.61 & -- \\
Par-Performer & 75.34 & 75.93 & 81.72 & 81.72 & -- \\
\bottomrule
\end{tabular}
}
\end{center}
\caption{Overall comparison of different attention mechanisms on ImageNet classification top-1 accuracy (\%), with input size 224. Tiny-4stage~/~4 means that the model has a comparable size with DeiT-Tiny, has 4 stages and uses patch size 4x4 in the initial pixel space. ``1,2,8,1'' are the numbers of attention blocks in each stage. ``Par-xformer" indicates multi-scale ViTs with multiple attention types: the first two stages utilize the ``xformer" efficient attention and the last two stages still use full attention. In the ``Trans2Det" columns, $\cmark$ indicates that the ImageNet-pre-trained model can be used to initialize detection backbones, $\xmark$ means not. $-$ means that it can be transferred, but the corresponding detection models consumes prohibitively large memory due to the need of using high resolution feature maps. SRA/32 downsizes key/value feature maps with the same schedule in PVT\cite{wang2021pyramid}, while SRA/64 downsizes more aggressively to make the memory managerable for downstream high-resolution tasks.}
\label{tab:eff_attns}
\end{table}

\noindent\textbf{Compare Vision Longformer with other attention mechanisms.} On the ImageNet classification task in Table~\ref{tab:eff_attns}, all efficient attention mechanisms above show a large performance gap from Vision Longformer. Linformer performs very competitively. Global attention and Performer have a similar performance with the DeiT model (72.2 for tiny and 79.8 for small). We use spatial reduction ratios $16,8,4,2$ from stage1 to stage4 for the multi-scale SRA model, which is different from the reduction ratios $8,4,2,1$ in PVT~\cite{wang2021pyramid}. This more aggressive spatial reduction makes the classification performance worse in Table~\ref{tab:eff_attns}, but makes the memory cost manageable when transfer to detection tasks for input image size $8000\times 1333$. For a more complete comparison of these models, including model parameters, FLOPs and memory usage, please refer to the Supplementary. 

\noindent\textbf{Why is Longformer better?} One possible reason is that the conv-like sparsity is a good inductive bias for vision transformers, compared with other attention mechanisms. This is supported by the visualization of the attention maps from pretrained DeiT models~\cite{touvron2020training}. Another explanation is that Vision Longformer keeps the key and value feature maps high resolution. However, low resolution-based attention mechanims like Linformer and SRA and pure global attention lose the high-resolution information in the key and value feature maps. 

\noindent\textbf{Mixed attention mechanisms (Partial X-former) for classification tasks.}
For classification tasks with $224\times 224$ image size as input, the feature map size at Stage3 in multi-scale ViTs is $14\times 14$. This is the same as the feature map size in ViT and DeiT, and best suits for full attention. A natural choice is to use efficient attention in the first two stages (with high-resolution feature map but with small number of blocks) and to use full attention in the last two stages. Multi-scale ViTs with this mixed attention mechanisms are called ``Parital X-former''. We also report these Partial X-formers' performance in Table~\ref{tab:eff_attns}. All these Partial X-formers perform well on ImageNet classification, with very little (even no) gap between Full Attention and Vision Longformer. These Partial X-forms achieve very good accuracy-efficiency performance for low-resolution classification tasks. We do not have ``Partial  ViL" for classification because ViL's window size is 15, and thus its attention mechanism in the last two stages is equivalent to the full attention.

%% file: detection.tex
\subsection{Transfer to High-resolution Vision Tasks}

Similar to the transfer-ability of ImageNet-pretrained CNN weights to downstream high-resolution tasks, such as object detection and segmentation, multi-scale Vision Longformer pretrained on ImageNet can be transferred to such high-resolution tasks, as we will show in Section~\ref{subsec:detexps}. 

However, Linformer is not transferable because the weights of the linear projection layer is specific to a resolution. The Partial X-formers and Multi-scale ViT with full attention are not transferable due to its prohibitively large memory usage after transferred to high-resolution tasks. In Table~\ref{tab:attention_maskrcnn}, we also show the superior performance of Vision Longformer over other attention mechanisms, on the object detection and segmentation tasks. 


%% file: classification_exps.tex
In this section, we show the final performance of Multi-scale Vision Longformer (short for \vil) on ImageNet classification in Section~\ref{subsec:class_exps} \& \ref{subsec:in21kexp} and downstream high-resolution detection tasks in Section~\ref{subsec:detexps}. We follow the DeiT training configuration for ImageNet classification training, and use the standard ``$\times 1$" and ``$\times 3$+MS" training schedules with the ``AdamW" optimizer for detection tasks. We refer to the Supplementary for detailed experimental settings.

\subsection{ImageNet Classification}
\label{subsec:class_exps}
Following DeiT\cite{touvron2020training} and PVT\cite{wang2021pyramid}, we build multi-scale ViLs with four different sizes, i.e., tiny, small, medium and base. 
The detailed model configuration is specified in Table~\ref{tab:model_arch}. We train multi-scale ViLs purely on ImageNet1K, following the setting in DeiT~\cite{touvron2020training}. 

In Table~\ref{tab:overall_comp_cls}, we report our results and compare with ResNets\cite{he2016deep}, ViT~\cite{dosovitskiy2020image}, DeiT~\cite{touvron2020training} and PVT~\cite{wang2021pyramid}. Our models outperform other models in the same scale by a large margin.
We again confirm that the relative positional bias (RPB) outperforms the absolute 2-D positional embedding (APE) on Vision Longformer. 
When compared with Swin Transformers~\cite{liu2021swin}, our models still performs better with fewer parameters. 

\begin{table}[ht]
\begin{center}
\footnotesize
\begin{tabular}{l@{\hspace{3pt}}|c@{\hspace{3pt}}|c|c}
\toprule
Model & \#Params (M) & GFLOPs & Top-1 (\%) \\
\midrule
R18 & 11.7 & 1.8 & 69.8 \\
DeiT-Tiny/16\cite{touvron2020training} & 5.7 & 1.3 & 72.2 \\
PVT-Tiny\cite{wang2021pyramid} & 13.2 & 1.9 & 75.1 \\
\rowcolor{Graylight} 
\vil-Tiny-APE & 6.7 & 1.3 & 76.3 \\
\rowcolor{Graylight} 
\vil-Tiny-RPB & 6.7 & 1.3 & 76.7 \\
\midrule
R50 & 25.6 & 4.1 & 78.5 \\
DeiT-Small/16\cite{touvron2020training} & 22.1 & 4.6 & 79.9 \\
PVT-Small\cite{wang2021pyramid} & 24.5 & 3.8 & 79.8 \\
Swin-Tiny\cite{liu2021swin} & 28 & 4.5 & 81.2 \\
\rowcolor{Graylight} 
\vil-Small-APE & 24.6 & 4.9 & 82.0 \\
\rowcolor{Graylight} 
\vil-Small-RPB & 24.6 & 4.9 & 82.4 \\
\midrule
R101 & 44.7 & 7.9 & 79.8 \\
PVT-Medium\cite{wang2021pyramid} & 44.2 & 6.7 & 81.2 \\
Swin-Small\cite{liu2021swin} & 50 & 8.7 & 83.2 \\
\rowcolor{Graylight} 
\vil-Medium-APE & 39.7 & 8.7 & 83.3 \\
\rowcolor{Graylight} 
\vil-Medium-RPB & 39.7 & 8.7 & 83.5 \\
\midrule
X101-64x4d & 83.5 & 15.6 & 81.5 \\
ViT-Base/16\cite{dosovitskiy2020image} & 86.6 & 17.6 & 77.9 \\
DeiT-Base/16\cite{touvron2020training} & 86.6 & 17.6 & 81.8 \\
PVT-Large\cite{wang2021pyramid} & 61.4 & 9.8 & 81.7 \\
Swin-Base\cite{liu2021swin} & 88 & 15.4 & 83.5 \\
\rowcolor{Graylight} 
\vil-Base-APE & 55.7 & 13.4 & 83.2 \\
\rowcolor{Graylight} 
\vil-Base-RPB & 55.7 & 13.4 & 83.7 \\
\bottomrule
\end{tabular}
\end{center}
\caption{Number of paramers, FLOPS and ImageNet accuracy. Trained on ImageNet-1K with image size 224. Our \vil~~models are highlighted with gray background.}
\label{tab:overall_comp_cls}
\end{table}

\input{imagenet22k_exps}

%% file: imagenet22k_exps.tex
\subsection{ImageNet-21K pretraining and ImageNet-1K finetuning}
\label{subsec:in21kexp}
When trained purely on ImageNet-1K, the performance gain from \vil-Medium to \vil-Base is very marginal. This is consistent with the observation in ViT~\cite{dosovitskiy2020image}: large pure transformer based models can be trained well only when training data is sufficient. 

Therefore, we conducted experiments in which \vil-Medium/Base models are first pre-trained on ImageNet-21k with image size $224^2$ and finetuned on ImageNet-1K with image size $384^2$. For ViT models on image size $384^2$, there are in total $24\times 24$ tokens with full attention. For \vil~models on image size $384^2$, we set the window sizes to be $(13, 17, 25, 25)$ from Stage1 to Stage4. Therefore, in the last two stages, the \vil~models' attention is still equivalent to full attention. 

As shown in In Table~\ref{tab:imagenet22k}, the performance gets significantly boosted after ImageNet-21K pretraining for both \vil~medium and base models. 
We want to point out that the performance of \vil-Medium model has surpassed that of ViT-Base/16, ViT-Large/16 and BiT-152x4-M, in the ImageNet-21K pretraining setting. The performance of \vil-Base models are even better. This shows the superior performance and parameter efficiency of \vil~models.

\begin{table}[ht]
\begin{center}
\resizebox{\linewidth}{!}{
\begin{tabular}{l@{\hspace{3pt}}|c@{\hspace{3pt}}|c|c|c|c}
\toprule
\multirow{2}{*}{Model} & \#Params & \multicolumn{2}{c|}{No IN-21K} & \multicolumn{2}{c}{After IN-21K} \\
& (M) & GFLOPs & Top-1 & GFLOPs & Top-1 \\
\midrule
ViT-Base/16\cite{dosovitskiy2020image} & 86.6 & 17.6 & 77.9 & 49.3 & 84.0\\
ViT-Large/16\cite{dosovitskiy2020image} & 307 & 61.6 & 76.5 & 191.1 & 85.2 \\
\midrule
BiT-152x4-M\cite{kolesnikov2019big} & 928 & 182 & 81.3 & 837 & 85.4 \\
\midrule
Swin-Base\cite{liu2021swin} & 88 & 15.4 & 83.5 & 47.1 & 86.4 \\
\midrule
\rowcolor{Graylight} 
\vil-Medium-RPB & 39.7 & 8.7 & 83.5 & 28.4 & 85.7 \\
\midrule
\rowcolor{Graylight} 
\vil-Base-RPB & 55.7 & 13.4 & 83.7 & 43.7 & 86.2 \\
\bottomrule
\end{tabular}
}
\end{center}
\caption{Trained purely on ImageNet-1K with image size 224 (No IN-21K). Pretained on ImageNet-21K with image size 224 and Finetuned on ImageNet-1K with image size 384 (After IN-21K), except BiT-M~\cite{kolesnikov2019big} fine-tuned with image size 480. Our \vil~~models are highlighted with gray background.}
\label{tab:imagenet22k}
\end{table}

%% file: detection_exps.tex
\subsection{Detection Tasks}
\label{subsec:detexps}

\input{detection_relative_all}

We apply our ViL to two representative object detection pipelines including RetinaNet~\cite{lin2017focal} and Mask-RCNN~\cite{he2017mask}. We follow the conventional setting to use our Vision Longformer as the backbone to generate feature maps for both detection pipelines. Similar to~\cite{wang2021pyramid}, we extract the features from all four scales and then feed them to the detection and/or instance segmentation head. To adapt the learned relative positional bias to the higher image resolution in detection, we perform bilinear interpolation on it prior to the training. In our experiments, all models are evaluated on COCO dataset~\cite{lin2014microsoft}, with 118k images for training and 5k images for evaluation. We report the results for both 1$\times$ and 3$\times$+MS training schedules, and compare them with two backbone architectures: ResNet~\cite{he2016deep} and PVT~\cite{he2017mask}. 

As shown in Table~\ref{tab:retinanet_comp_det}, our ViL achieves significantly better performance than the ResNet and PVT architecture. The improvements are uniform over all model sizes (tiny, small, medium, base) and over all object scales ($AP_S$, $AP_M$, $AP_L$). The improvement is so large that \vil-Tiny with ``3x+MS" schedule already outperforms the ResNeXt101-64x4d and the PVT-Large models. A similar trend is observed with the Mask R-CNN pipeline. As shown in Table~\ref{tab:maskrcnn_comp_det}, our ViL backbone significantly surpasses ResNet and PVT baselines on both object detection and instance segmentation. When compared with the concurrent Swin Transformer~\cite{liu2021swin}, our model also outperforms it with fewer parameter and FLOPs. More specifically, our \vil-Small achieves 47.1 $AP^b$ with 45M parameters, while Swin-Tiny achieves 46.0 $AP^b$ with 48M parameters. These consistent and significant improvements with both RetinaNet and Mask R-CNN demonstrate the promise of our proposed ViL when using it as the image encoder for high-resolution dense object detection tasks.

%% file: detection_relative_all.tex
\begin{table*}[ht]
\begin{center}
\resizebox{\linewidth}{!}{
\begin{tabular}{l@{\hspace{3pt}}|c@{\hspace{3pt}}|c|c|c|c|c|c|c|c|c|c|c|c|c}
\toprule
\multirow{2}{*}{Backbone} & \#Params & FLOPs & \multicolumn{6}{c|}{RetinaNet 1x schedule} & \multicolumn{6}{c}{RetinaNet 3x + MS schedule}\\
& (M) & (G) & $AP$ & $AP_{50}$ & $AP_{75}$ & $AP_{S}$ & $AP_{M}$ & $AP_{L}$ & $AP$ & $AP_{50}$ & $AP_{75}$ & $AP_{S}$ & $AP_{M}$ & $AP_{L}$ \\
\midrule
ResNet18 & 21.3 & 190.33 & 31.8 & 49.6 & 33.6 & 16.3 & 34.3 & 43.2 & 35.4 & 53.9 & 37.6 & 19.5 & 38.2 & 46.8 \\
PVT-Tiny\cite{wang2021pyramid} & 23.0 & n/a & 36.7 & 56.9 & 38.9 & 22.6 & 38.8 & 50.0 & 39.4 & 59.8 & 42.0 & 25.5 & 42.0 & 52.1 \\
\rowcolor{Graylight} 
\vil-Tiny-RPB & 16.64 & 182.7 & 40.8 & 61.3 & 43.6 & 26.7 & 44.9 & 53.6 & 43.6 & 64.4 & 46.1 & 28.1 & 47.5 & 56.7 \\
\midrule
ResNet50 & 37.7 & 239.32 & 36.3 & 55.3 & 38.6 & 19.3 & 40.0 & 48.8 & 39.0 & 58.4 & 41.8 & 22.4 & 42.8 & 51.6 \\
PVT-Small\cite{wang2021pyramid} & 34.2 & n/a & 40.4 & 61.3 & 43.0 & 25.0 & 42.9 & 55.7 & 42.2 & 62.7 & 45.0 & 26.2 & 45.2 & 57.2\\
\rowcolor{Graylight} 
\vil-Small-RPB & 35.68 & 254.8 & 44.2 & 65.2 & 47.6 & 28.8 & 48.0 & 57.8 & 45.9 & 66.6 & 49.0 & 30.9 & 49.3 & 59.9 \\
\midrule
ResNet101 & 56.7 & 319.07 & 38.5 & 57.8 & 41.2 & 21.4 & 42.6 & 51.1 & 40.9 & 60.1 & 44.0 & 23.7 & 45.0 & 53.8 \\
ResNeXt101-32x4d & 56.4 & 319.07 & 39.9 & 59.6 & 42.7 & 22.3 & 44.2 & 52.5 & 41.4 & 61.0 & 44.3 & 23.9 & 45.5 & 53.7 \\
PVT-Medium\cite{wang2021pyramid} & 53.9 & n/a & 41.9 & 63.1 & 44.3 & 25.0 & 44.9 & 57.6 & 43.2 & 63.8 & 46.1 & 27.3 & 46.3 & 58.9 \\
\rowcolor{Graylight} 
\vil-Medium-RPB & 50.77 & 330.4 & 46.8 & 68.1 & 50.0 & 31.4 & 50.8 & 60.8 & 47.9 & 68.8 & 51.3 & 32.4 & 51.9 & 61.8 \\
\midrule
ResNeXt101-64x4d & 95.5 & 483.59 & 41.0 & 60.9 & 44.0 & 23.9 & 45.2 & 54.0 & 41.8 & 61.5 & 44.4 & 25.2 & 45.4 & 54.6 \\
PVT-Large\cite{wang2021pyramid} & 71.1 & n/a & 42.6 & 63.7 & 45.4 & 25.8 & 46.0 & 58.4 & 43.4 & 63.6 & 46.1 & 26.1 & 46.0 & 59.5 \\
\rowcolor{Graylight} 
\vil-Base-RPB & 66.74 & 420.9 & 47.8 & 69.2 & 51.4 & 32.4 & 52.3 & 61.8 & 48.6 & 69.4 & 52.2 & 34.1 & 52.5 & 61.9 \\
\bottomrule
\end{tabular}
}
\end{center}
\vspace{-2mm}
\caption{Object detection performance on the COCO val2017 with RetinaNet. The FLOPs (G) are measured at resolution $800\times 1333$, and FLOPs for PVT architecture are not available. Our \vil-Tiny and \vil-Small models are pre-trained on ImageNet-1K, our \vil-Medium and \vil-Base models are pre-trained on ImageNet-21k. \vil~results are highlighted with gray background.}
\label{tab:retinanet_comp_det}
\vspace{-2mm}
\end{table*}

\begin{table*}[ht]
\begin{center}
\resizebox{\linewidth}{!}{
\begin{tabular}{l@{\hspace{3pt}}|c@{\hspace{3pt}}|c|c|c|c|c|c|c|c|c|c|c|c|c}
\toprule
\multirow{2}{*}{Backbone} & \#Params & FLOPs & \multicolumn{6}{c|}{Mask R-CNN 1x schedule} & \multicolumn{6}{c}{Mask R-CNN 3x + MS schedule}\\
 & (M) & (G) & $AP^b$ & $AP^b_{50}$ & $AP^b_{75}$ & $AP^m$ & $AP^m_{50}$ & $AP^m_{75}$ & $AP^b$ & $AP^b_{50}$ & $AP^b_{75}$ & $AP^m$ & $AP^m_{50}$ & $AP^m_{75}$ \\
\midrule
ResNet18 & 31.2 & 131.03 & 34.0 & 54.0 & 36.7 & 31.2 & 51.0 & 32.7 & 36.9 & 57.1 & 40.0 & 33.6 & 53.9 & 35.7\\
PVT-Tiny\cite{wang2021pyramid} & 32.9 & n/a & 36.7 & 59.2 & 39.3 & 35.1 & 56.7 & 37.3 & 39.8 & 62.2 & 43.0 & 37.4 & 59.3 & 39.9 \\
\rowcolor{Graylight} 
\vil-Tiny-RPB & 26.9 & 145.6 & 41.4 & 63.5 & 45.0 & 38.1 & 60.3 & 40.8 & 44.2 & 66.4 & 48.2 & 40.6 & 63.2 & 44.0 \\
\midrule
ResNet50 & 44.2 & 180.0 & 38.0 & 58.6 & 41.4 & 34.4 & 55.1 & 36.7 & 41.0 & 61.7 & 44.9 & 37.1 & 58.4 & 40.1 \\
PVT-Small\cite{wang2021pyramid} & 44.1 & n/a & 40.4 & 62.9 & 43.8 & 37.8 & 60.1 & 40.3 & 43.0 & 65.3 & 46.9 & 39.9 & 62.5 & 42.8 \\
Swin-Tiny\cite{liu2021swin} & 48 & 267 & -- & -- & -- & -- & -- & -- & 46.0 & 68.1 & 50.3 & 41.6 & 65.1 & 44.9 \\
\rowcolor{Graylight} 
\vil-Small-RPB & 45.0 & 218.3 & 44.9 & 67.1 & 49.3 & 41.0 & 64.2 & 44.1 & 47.1 & 68.7 & 51.5 & 42.7 & 65.9 & 46.2 \\
\midrule
ResNet101 & 63.2 & 259.77  & 40.4 & 61.1 & 44.2 & 36.4 & 57.7 & 38.8 & 42.8 & 63.2 & 47.1 & 38.5 & 60.1 & 41.3 \\
ResNeXt101-32x4d & 62.8 & 259.77 & 41.9 & 62.5 & 45.9 & 37.5 & 59.4 & 40.2 & 44.0 & 64.4 & 48.0 & 39.2 & 61.4 & 41.9 \\
PVT-Medium\cite{wang2021pyramid} & 63.9 & n/a & 42.0 & 64.4 & 45.6 & 39.0 & 61.6 & 42.1 & 44.2 & 66.0 & 48.2 & 40.5 & 63.1 & 43.5 \\
Swin-Small\cite{liu2021swin} & 69 & 359 & -- & -- & -- & -- & -- & -- & 48.5 & 70.2 & 53.5 & 43.3 & 67.3 & 46.6 \\
\rowcolor{Graylight} 
\vil-Medium-RPB & 60.1 & 293.8 & 47.6 & 69.8 & 52.1 & 43.0 & 66.9 & 46.6 & 48.9 & 70.3 & 54.0 & 44.2 & 67.9 & 47.7 \\
\midrule
ResNeXt101-64x4d & 101.9 & 424.29 & 42.8 & 63.8 & 47.3 & 38.4 & 60.6 & 41.3 & 44.4 & 64.9 & 48.8 & 39.7 & 61.9 & 42.6 \\
PVT-Large\cite{wang2021pyramid} & 81.0 & n/a & 42.9 & 65.0 & 46.6 & 39.5 & 61.9 & 42.5 & 44.5 & 66.0 & 48.3 & 40.7 & 63.4 & 43.7 \\
\rowcolor{Graylight} 
\vil-Base-RPB & 76.1 & 384.4 & 48.6 & 70.5 & 53.4 & 43.6 & 67.6 & 47.1 & 49.6 & 70.7 & 54.6 & 44.5 & 68.3 & 48.0 \\
\bottomrule
\end{tabular}
}
\end{center}
\vspace{-2mm}
\caption{Object detection and instance segmentation performance on the COCO val2017 with Mask R-CNN. The FLOPs (G) are measured at resolution $800\times 1333$, and FLOPs for PVT architecture are not available. Our \vil-Tiny and \vil-Small models are pre-trained on ImageNet-1K, our \vil-Medium and \vil-Base models are pre-trained on ImageNet-21k. \vil~results are highlighted with gray background.}
\label{tab:maskrcnn_comp_det}
\vspace{-3mm}
\end{table*}

%% file: ablation_studies.tex
\vspace{-2mm}
\subsection{Ablation Study for Detection Tasks}
\label{subsec:ablation_classification}

\begin{table}[ht]
\begin{center}
\resizebox{\linewidth}{!}{
\footnotesize
\begin{tabular}{l@{\hspace{3pt}}|c@{\hspace{3pt}}|c|c|c|c}
\toprule
Attention & \shortstack{\#Params\\(M)} & $AP^b$ & $AP^m$ & \shortstack{FLOPs\\(G)} & \shortstack{Memory\\(G)}\\
\midrule
SRA/64~\cite{wang2021pyramid} & 73.3 & 36.4 & 34.6 & 224.1 & 7.1 \\
SRA/32~\cite{wang2021pyramid} & 51.5 & 39.9 & 37.3 & 268.3 & 13.6 \\
Par-SRA/32  & 46.8 & 42.4 & 39.0 & 352.1 & 22.6 \\
\hline
Global & 45.2 & 34.8 & 33.4 & 226.4 & 7.6 \\
Par-Global  & 45.1 & 42.5 & 39.2 & 326.5 & 20.1  \\
\hline
Performer & 45.0 & 36.1 & 34.3 & 251.5 & 8.4 \\
Par-Performer  & 45.0 & 42.3 & 39.1 & 343.7 & 20.0  \\
\hline
ViL & 45.0 & 42.9 & 39.6 & 218.3 & 7.4 \\
Par-ViL  & 45.0 & 43.3 & 39.8  & 326.8 & 19.5 \\
\bottomrule
\end{tabular}
}
\end{center}
\vspace{-1mm}
\caption{Comparing different efficient attention mechanisms for object detection with Mask R-CNN. All use small model size and absolute 2-D positional embedding (APE) for fair comparison. Run-time memory cost when training each model is also reported.}
\label{tab:attention_maskrcnn}
\end{table}

\noindent\textbf{Compare with other efficient attention mechanisms}. Similar to Sec~\ref{subsec:ablation_classification}, we study SRA~\cite{wang2021pyramid}, Global Transformer and Performer and their corresponding partial version with Mask R-CNN pipeline (trained with the 1$\times$ schedule). As we can see in Table~\ref{tab:attention_maskrcnn}, when efficient attention mechanisms are used in all stages, ViL achieves much better performance than the other three mechanisms. Specifically, our ViL achieves 42.9 $AP^b$ while the other three are all around 36.0 $AP^b$. 
When efficient attention mechanisms are only used in the first two stages (Par-Xformer), the gaps between different mechanisms shrink to around 1.0 point while our ViL still outperform all others. 
Moreover, the \vil~model outperforms the partial models of all other attention mechanisms and has a very small gap (0.4 $AP^b$) from the Partial-\vil~model. 
These results show that the ``local attention + global memory" mechanism in Vision Longformer can retain the good performance of the full attention mechanism in ViT, and that it is a clear better choice than other efficient attention mechanisms for high-resolution vision tasks.

\begin{figure}[t!]
\includegraphics[width=0.235\textwidth]{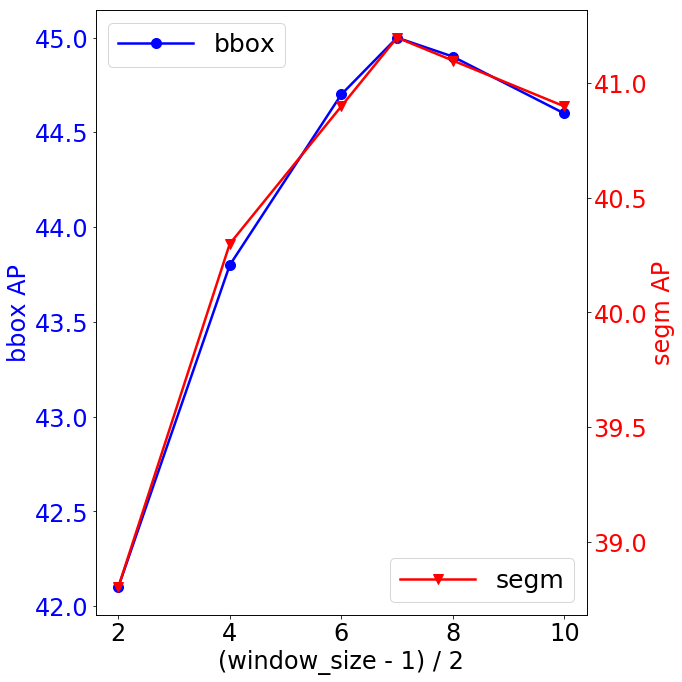}
\includegraphics[width=0.235\textwidth]{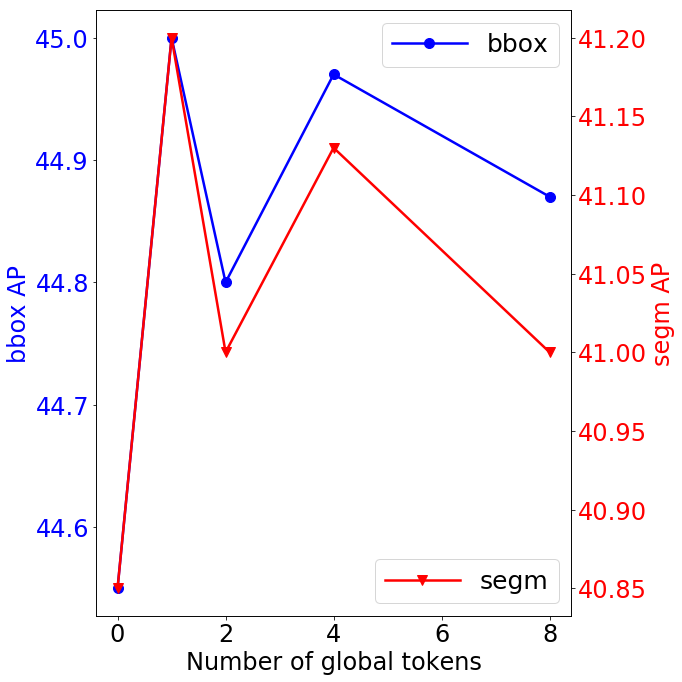}
\caption{Effects of window size (Left) and number of global tokens (Right) in Vision Longformer for object detection with Mask R-CNN. All use the same ImageNet1K pre-trained checkpoint (\vil-Small-RPB in Table~\ref{tab:overall_comp_cls}).}
\label{fig:vil_winsize_nglo}
\vspace{-2mm}
\end{figure}

\noindent\textbf{The effects of window size and number of global tokens} are not obvious in ImageNet classification, as long as the last two stages use full attention. For different window sizes in $[9, 15, 21]$ and different number of global tokens in $[0, 1, 2, 4, 8]$, the final top-1 accuracy differs by at most 0.2 for \vil-Small models. Meanwhile, their effects are significant in high-resolution tasks, where \vil~models use local attention in all stages. In Figure~\ref{fig:vil_winsize_nglo}, we report their effects in COCO object detection with Mask R-CNN. We notice that the window size plays a crucial role and the default window size 15 gives the best performance. Smaller window sizes lead to serious performance drop. As shown in Figure~\ref{fig:vil_winsize_nglo} (Right), as long as there is one global token, adding more global tokens does not improve the performance any more. 

%% file: conclusion.tex
\section{Conclusions}
\vspace{-2mm}
In this paper, we have presented a new Vision Transformer (ViT) architecture \emph{Multi-Scale Vision Longformer} to address the computational and memory efficiency that prevents the vanilla ViT model from applying to vision tasks requiring high-resolution feature maps. We mainly developed two techniques: 1) a multi-scale model structure designed for Transformers to provide image encoding at multiple scales with manageable computational cost, and 2) an efficient 2-D attention mechanism of Vision Longformer for achieving a linear complexity w.r.t. the number of input tokens. The architecture design and the efficient attention mechanism are validated with comprehensive ablation studies. Our experimental results show that the new ViT architecture effectively addresses the computational and memory efficiency problem and outperforms several strong baselines on image classification and object detection.

%% file: app_settings.tex
\section{Settings}
\label{app:settings}
\subsection{Model configurations}
\label{app:model_archs}
We listed the model configuration of all models used in this paper in Table~\ref{tab:model_arch_full}. We do not specify the attention mechanism here, because the model configuration is the same for all attention mechanisms and the attention-specific parameters are specified in Table~\ref{tab:attention_param_full}.

\begin{table}[ht]
\begin{center}
\resizebox{\linewidth}{!}{
\begin{tabular}{l@{\hspace{1.5pt}}|c@{\hspace{3pt}}|c@{\hspace{3pt}}|c@{\hspace{3pt}}|c@{\hspace{3pt}}}
\toprule
\multirow{2}{*}{Model} & Stage1 & Stage2 & Stage3 & Stage4 \\
 & n,p,h,d & n,p,h,d & n,p,h,d & n,p,h,d \\
\midrule
Tiny & 1,4,1,48 & 1,2,3,96 & 9,2,3,192 & 1,2,6,384 \\
\hline
Small & 1,4,3,96 & 2,2,3,192 & 8,2,6,384 & 1,2,12,768 \\
\hline
Medium & 1,4,3,96 & 4,2,3,192 & 16,2,6,384 & 1,2,12,768 \\
\hline
Base & 1,4,3,96 & 8,2,3,192 & 24,2,6,384 & 1,2,12,768 \\
\midrule
Tiny 1-10-1 & \multicolumn{2}{c|}{1,8,3,96} & 10,2,3,192 & 1,2,6,384 \\
Tiny 2-9-1 & \multicolumn{2}{c|}{2,8,3,96} & 9,2,3,192 & 1,2,6,384 \\
Tiny 1-9-2 & \multicolumn{2}{c|}{1,8,3,96} & 9,2,3,192 & 2,2,6,384 \\
Tiny 2-8-2 & \multicolumn{2}{c|}{2,8,3,96} & 8,2,3,192 & 2,2,6,384 \\
Tiny 1-1-9-1 & 1,4,1,48 & 1,2,3,96 & 9,2,3,192 & 1,2,6,384 \\
Tiny 1-2-8-1 & 1,4,1,48 & 2,2,3,96 & 8,2,3,192 & 1,2,6,384 \\
\hline
Small 1-10-1 & \multicolumn{2}{c|}{1,8,3,192} & 10,2,6,384 & 1,2,12,768 \\
Small 2-9-1 & \multicolumn{2}{c|}{2,8,3,192} & 9,2,6,384 & 1,2,12,768 \\
Small 1-9-2 & \multicolumn{2}{c|}{1,8,3,192} & 9,2,6,384 & 2,2,12,768 \\
Small 2-8-2 & \multicolumn{2}{c|}{2,8,3,192} & 8,2,6,384 & 2,2,12,768 \\
Small 1-1-9-1 & 1,4,3,96 & 1,2,3,192 & 9,2,6,384 & 1,2,12,768 \\
Small 1-2-8-1 & 1,4,3,96 & 2,2,3,192 & 8,2,6,384 & 1,2,12,768 \\
\bottomrule
\end{tabular}
}
\end{center}
\caption{Model architecture for multi-scale stacked ViTs. Architecture parameters for each E-ViT module $\text{E-ViT}(a\times n/p~;h,d)$: number of attention blocks $n$, input patch size $p$, number of heads $h$ and hidden dimension $d$. See the meaning of these parameters in Figure~\ref{fig:ms_vit} (Bottom).}
\label{tab:model_arch_full}
\end{table}


\subsection{Experimental settings}
\label{app:exps}

\begin{table*}[ht]
\begin{center}
\resizebox{\linewidth}{!}{
\begin{tabular}{cccccccc}
\toprule
Model & Dataset & Epoch & Base Lr & LR decay & Weight decay & Drop Path & Batch size \\
\hline
MsViT-Tiny & ImageNet & 300 & 1e-3 & cosine & 0.1 & 0.1 & 1024\\
MsViT-Small & ImageNet & 300 & 1e-3 & cosine & 0.1 & 0.1 & 1024\\
MsViT-Meidum & ImageNet & 300 & 8e-4 & cosine & 0.1 & 0.1 & 1024\\
MsViT-Base & ImageNet & 150 & 8e-4 & cosine & 0.1 & 0.1 & 1024\\
\midrule
MsViT-Meidum & ImageNet-21k & 90 & 5e-4 & cosine & 0.1 & 0.1 & 1024\\
MsViT-Base & ImageNet-21k & 90 & 5e-4 & cosine & 0.1 & 0.1 & 1024\\
\midrule
MsViT-Meidum & ImageNet-384 & 10 & [2, 4]*e-2 & cosine & 0. & 0.1 & 512\\
MsViT-Base & ImageNet-384 & 10 & [2, 4]*e-2 & cosine & 0. & 0.1 & 512\\
\midrule
Model & Dataset & iterations & Base Lr & LR decay & Weight decay & Drop Path & Batch size \\
\hline
MsViT-Tiny-1x & COCO & 60k-80k-90k & 1e-4 & multi-step & 0.05 & [0.05, 0.1] & 16\\
MsViT-Small-1x & COCO & 60k-80k-90k & 1e-4 & multi-step & 0.05 & [0.1, 0.2] & 16\\
MsViT-Meidum-D-1x & COCO & 60k-80k-90k & 1e-4 & multi-step & 0.05 & [0.2, 0.3] & 16\\
MsViT-Base-D-1x & COCO & 60k-80k-90k & 8e-5 & multi-step & 0.05 & [0.2, 0.3] & 16\\
\hline
MsViT-Tiny-3x+ms & COCO & 180k-240k-270k & 1e-4 & multi-step & 0.05 & [0.05, 0.1] & 16\\
MsViT-Small-3x+ms & COCO & 180k-240k-270k & 1e-4 & multi-step & 0.05 & [0.1, 0.2] & 16\\
MsViT-Meidum-D-3x+ms & COCO & 180k-240k-270k & 1e-4 & multi-step & 0.05 & [0.2, 0.3] & 16\\
MsViT-Base-D-3x+ms & COCO & 180k-240k-270k & 8e-5 & multi-step & 0.05 & [0.2, 0.3] & 16\\
\bottomrule
\end{tabular}
}
\end{center}
\caption{Hyperparameters for training. We use MsViT to represent the multi-scale vision transformers with different kinds of attention mechanisms, including our Vision Longformer (\vil). For the experiments trained on COCO, MsViT is combined with the Retinanet or Mask R-CNN. The training configs for Retinanet or Mask R-CNN are the same, and we still use MsViT for their unified short name. We do not apply gradient clipping for all ImageNet classification training and apply gradient clipping at global norm $1$ for COCO object detection/segmentation. We use AdamW for all our experiments, except that we use SGD with momentum $0.9$ for the ImageNet-384 fine-tuning experiments.}
\label{tab:exp_config}
\end{table*}

Table~\ref{tab:exp_config} summarizes our training setups for our different models. 

For the ImageNet classification task, our setting mainly follow that in DeiT~\cite{touvron2020training}. For example, we do not use dropout but use random path. We use all data augmentations in DeiT~\cite{touvron2020training}, except that we apply Repeated Augmentation only on Medium and Base models. When fine-tuning from a ImageNet-21K pretrained checkpoint, we mainly follow the practice of ViT~\cite{dosovitskiy2020image}, train on image size $384\times 384$, use SGD with momentum $0.9$, use no weight decay, and use only random cropping for data augmentation. 

For COCO object detection/segmentation tasks, we follow the standard ``$1\times$" and ``$3\times+\text{MS}$" schedules. We only change the optimizer from SGD to AdamW and search for good initial learning rate and weight decay. For the ``$1\times$" schedule, the input image scale is fixed to be $(800,1333)$ for the min and max sizes, respectively. For the ``$3\times+\text{MS}$" schedule, the input image is randomly resized to have min size in $\{640, 672, 704, 736, 768, 800\}$. We found that there is obvious over-fitting in Training \vil-Medium and \vil-Base models on COCO, mainly because that these two models are relatively large but they are only pretrained on ImageNet. Therefore, we are taking the best checkpoint (one epoch per checkpoint) along the training trajectory to report the performance. 

%% file: appendix.tex
\section{More experimental results}



\subsection{Ablation study on the architecture design of multi-scale Vision Longformer}
In this section, we present two ablation studies on the model architecture of multi-scale Vision Longformer.

\noindent\textbf{Ablation of the effects of LayerNorm and 2-D positional embedding in the patch embedding.}
In Table~\ref{tab:flatvsmultiscale}, we show that our flat model $\text{E-ViT}(\text{full}\times 12/16)$, which only differs from the standard ViT/DeiT model by an newly-added LayerNorm after the patch embedding and the 2-D positional embedding, has better performance than the standard ViT/DeiT model. In Table~\ref{tab:layernorm_2dposition}, we show that this better performance comes from the newly-added LayerNorm.
\begin{table}[ht]
\begin{center}
\begin{tabular}{l@{\hspace{3pt}}|cc|cc}
\toprule
\multirow{2}{*}{Model} & \multicolumn{2}{c|}{Tiny} & \multicolumn{2}{c}{Small} \\
& CLS & Ave Pool & CLS & Ave Pool \\
\midrule
DeiT/16\cite{touvron2020training} & 72.2 & - & 79.8 & - \\
+Layernorm & 72.91 & 73.36 & 80.33 & 80.32 \\
+2D Pos & 73.21 & 73.09 & 80.44 & 80.75 \\
\bottomrule
\end{tabular}
\end{center}
\caption{Ablation of the effects of LayerNorm and 2-D positional embedding in the patch embedding of the E-ViT module, with ImageNet Top-1 accuracy. The improvement over DeiT~\cite{touvron2020training} comes from the added LayerNorm. The 2-D positional embedding is mainly for saving parameters for high-resolution feature maps. The column names of ``CLS" and ``Ave Pool" indicate how the image feature is obtained for the linear classification head.}
\label{tab:layernorm_2dposition}
\end{table}

\noindent\textbf{Feature from the CLS token or from average pooling?} As shown in Table~\ref{tab:cls_avepool}, for ViL models that has only one attention block in the last stage (\vil~1-2-8-1), the average pooled feature from all tokens works better than the feature of the CLS token. However, when there are more than 2 attention blocks in the last stage (\vil~1-1-8-2), the difference between these two features disappears. The \vil~1-1-8-2 model has better performance than the \vil~1-2-8-1 model because it has more trainable parameters. 
\begin{table}[ht] 
\begin{center}
\begin{tabular}{l@{\hspace{3pt}}|cc|cc}
\toprule
\multirow{2}{*}{Model} & \multicolumn{2}{c|}{Tiny} & \multicolumn{2}{c}{Small} \\
& CLS & Ave Pool & CLS & Ave Pool \\
\hline
\vil-1,2,8,1-APE & 75.72 & 75.98 & 81.65 & 81.99 \\
\vil-1,1,8,2-APE & 76.18 & 76.25 & 82.12 & 82.08 \\
\bottomrule
\end{tabular}
\end{center}
\caption{For ViL models that has only one attention block in the last stage (\vil~1-2-8-1), the average pooled feature from all tokens works better than the feature of the CLS token. When there are more than 2 attention blocks in the last stage (\vil~1-1-8-2), the difference between these two features disappears.}
\label{tab:cls_avepool}
\end{table}

\subsection{A comprehensive comparison of different attention mechanisms on ImageNet classification}
We compare different attention mechanisms with different model sizes and architectures in Table~\ref{tab:attention_overall} and Table~\ref{tab:attention_overall_timespace}. In Table~\ref{tab:attention_overall}, we show their performance on ImageNet-1K classification problem, measured by Top-1 accuracy. In Table~\ref{tab:attention_overall_timespace}, we show their number of parameters and FLOPs. We would like to comment that FLOPs is just a theoretical estimation of computation complexity, and it may not fit well the space/time cost in practice.

\input{master_table}

%% file: master_table.tex
\begin{table*}[!ht]
\begin{center}
\resizebox{\linewidth}{!}{
\begin{tabular}{c@{\hspace{3pt}}|c@{\hspace{5pt}}c@{\hspace{3pt}}|c@{\hspace{5pt}}c|c@{\hspace{6pt}}c@{\hspace{6pt}}|c@{\hspace{6pt}}c}
\toprule
Flat Models & \multicolumn{4}{c|}{Tiny} & \multicolumn{4}{c}{Small} \\ 
\midrule
DeiT~/~16~\cite{touvron2020training} & \multicolumn{4}{c|}{72.2} & \multicolumn{4}{c}{79.8} \\
$\text{E-ViT}(\text{full}\times 12/16)$ & \multicolumn{4}{c|}{73.21 (CLS) / 73.09 (AVG)} & \multicolumn{4}{c}{80.44 (CLS) / 80.75 (AVG)}  \\
\midrule
Multi-scale & \multicolumn{2}{c|}{Tiny-3stage~/~8} & \multicolumn{2}{c|}{Tiny-4stage~/~4} & \multicolumn{2}{c|}{Small-3stage~/~8} & \multicolumn{2}{c}{Small-4stage~/~4} \\ 
Models & 1-10-1 & 2-9-1 & 1-1-9-1 & 1-2-8-1 & 1-10-1 & 2-9-1 & 1-1-9-1 & 1-2-8-1 \\ \midrule
Full Attention & $75.86$ & $75.79$ & $76.06$ & 75.60 & $81.66$ & $81.73$ & $81.93^*$ & $81.91^*$  \\
\hline
Vision Longformer & 75.63\tiny{$\pm$0.23} & 75.88\tiny{$\pm$0.08} & 76.18\tiny{$\pm$0.12} & 75.98\tiny{$\pm$0.10} & 81.57 & 81.81 & 81.79 & 81.99 \\
\hline
Linformer~\cite{wang2020linformer} & 74.54 & 74.72 & 74.71 & 74.74 & 80.76 & 80.99 & 81.19 & 80.98  \\
Partial Linformer & 75.64 & 75.82 & 75.56 & 75.33 & 81.66 & 81.63 & 81.66 & 81.79 \\
\hline
SRA/64~\cite{wang2021pyramid} & 68.71 & 68.84 & 69.08 & 68.78 & 75.9 & 76.18 & 76.35 & 76.37  \\
SRA/32~\cite{wang2021pyramid} & 73.16 & 73.46 & 73.22 & 73.2 & 79.82 & 79.8 & 79.96 & 79.9 \\
Partial SRA/32 & 75.17 & 75.8 & 75.2 & 75.26 & 81.63 & 81.59 & 81.62 & 81.61  \\
\hline
Global & 70.93 & 71.62 & 71.52 & 72.00 & 79.04 & 79.08 & 79.17 & 78.97  \\
Partial Global & 75.55 & 75.61 & 75.32 & 75.4 & 81.39 & 81.42 & 81.6 & 81.45  \\
\hline
Performer & 71.28 & 71.87 & 71.12 & 73.09 & 78.17 & 78.58 & 78.81 & 78.72 \\
Partial Performer & 75.65 & 75.74 & 75.34 & 75.93 & 81.59 & 81.86 & 81.72 & 81.72 \\
\bottomrule
\end{tabular}
}
\end{center}
\caption{Overall comparison in ImageNet top-1 accuracy, with input size 224. Tiny-4stage~/~4 means that the model has comparable size with DeiT-Tiny, has 4 stages and uses patch size 4x4 in the initial pixel space. 1-2-8-1 means that the model contains 4 stages, each stage has 1/2/8/1 MSA-FFN blocks, respectively. *Partial* means that the last two stages, which contain most of the attention blocks, still use full attention. Vision Longformer does not have *Partial* version because its window size is set as 15 (comparable with the ViT(DeiT)/16 feature map size 14), and its attention mechanism in the last two stages is equivalent to full attention. * indicates that the training batch size is 256 (with learning rate linearly scaled down), different from all other experiments with batch size 1024 in this table.}
\label{tab:attention_overall}
\end{table*}

\begin{table*}[!ht]
\begin{center}
\resizebox{\linewidth}{!}{
\begin{tabular}{c@{\hspace{3pt}}|c@{\hspace{5pt}}c@{\hspace{3pt}}|c@{\hspace{5pt}}c|c@{\hspace{6pt}}c@{\hspace{6pt}}|c@{\hspace{6pt}}c}
\toprule
Flat Models & \multicolumn{4}{c|}{Tiny} & \multicolumn{4}{c}{Small} \\ 
\midrule
DeiT~/~16~\cite{touvron2020training} & \multicolumn{4}{c|}{5.7 (M) parameters, 1.3 GFLOPs} & \multicolumn{4}{c}{22.1 (M) parameters, 4.6 GFLOPs} \\
$\text{E-ViT}(\text{full}\times 12/16)$ & \multicolumn{4}{c|}{5.7 (M) parameters, 1.3 GFLOPs} & \multicolumn{4}{c}{22.1 (M) parameters, 4.6 GFLOPs}  \\
\midrule
Multi-scale & \multicolumn{2}{c|}{Tiny-3stage~/~8} & \multicolumn{2}{c|}{Tiny-4stage~/~4} & \multicolumn{2}{c|}{Small-3stage~/~8} & \multicolumn{2}{c}{Small-4stage~/~4} \\ 
Models & 1-10-1 & 2-9-1 & 1-1-9-1 & 1-2-8-1 & 1-10-1 & 2-9-1 & 1-1-9-1 & 1-2-8-1 \\ \midrule
Full Attention & 7.1, 1.35 & 6.8, 1.45 & 6.7, 2.29 & 6.4, 2.39 & 27.6, 4.84 & 26.3, 5.05 & 26.0, 6.74 & 24.6, 6.95  \\
\hline
Vision Longformer& 7.1, 1.27 & 6.8, 1.29 & 6.7, 1.33 & 6.4, 1.35 & 27.6, 4.67 & 26.3, 4.71 & 26.0, 4.82 & 24.6, 4.86 \\
\hline
Linformer~\cite{wang2020linformer} & 7.8, 1.57 & 7.7, 1.6 & 8.2, 1.69 & 8.0, 1.73 & 28.3, 5.27 & 27.1, 5.35 & 27.4, 5.55 & 26.3, 5.62  \\
Partial Linformer & 7.3, 1.31 & 7.2, 1.37 & 7.7, 1.46 & 7.6, 1.52 & 27.8, 4.76 & 26.7, 4.88 & 27.0, 5.08 & 25.8, 5.21 \\
\hline
SRA/64~\cite{wang2021pyramid} & 14.2, 0.99 & 13.9, 0.99 & 13.8, 1.0 & 13.5, 1.0 & 55.9, 3.92 & 54.6, 3.92 & 54.3, 3.97 & 52.9, 3.97  \\
SRA/32~\cite{wang2021pyramid} & 8.7, 1.09 & 8.4, 1.09 & 8.3, 1.1 & 8.0, 1.1 & 34.1, 4.23 & 32.7, 4.23 & 32.5, 4.28 & 31.1, 4.28 \\
Partial SRA/32 & 7.3, 1.23 & 7.1, 1.22 & 7.0, 1.24 & 6.8, 1.22 & 28.2, 4.6 & 27.4, 4.56 & 27.1, 4.61 & 26.4, 4.57  \\
\hline
Global & 7.2, 1.69 & 6.9, 1.7 & 6.8, 1.75 & 6.5, 1.76 & 27.8, 6.65 & 26.4, 6.67 & 26.2, 6.76 & 24.9, 6.78  \\
Partial Global & 7.2, 1.6 & 6.9, 1.62 & 6.8, 1.68 & 6.5, 1.7 & 27.8, 6.07 & 26.4, 6.14 & 26.2, 6.23 & 24.9, 6.3  \\
\hline
Performer & 7.3, 1.81 & 7.0, 1.86 & 6.9, 1.99 & 6.6, 2.05 & 27.8, 5.75 & 26.5, 5.87 & 26.2, 6.14 & 24.8, 6.26 \\
Partial Performer & 7.1, 1.35 & 6.8, 1.45 & 6.7, 1.57 & 6.4, 1.67 & 27.6, 4.84 & 26.3, 5.04 & 26.0, 5.31 & 24.7, 5.52 \\
\bottomrule
\end{tabular}
}
\end{center}
\caption{Overall comparison in number of parameters (M) and GFLOPs, with input size 224. Tiny-4stage~/~4 means that the model has comparable size with DeiT-Tiny, has 4 stages and uses patch size 4x4 in the initial pixel space. 1-2-8-1 means that the model contains 4 stages, each stage has 1/2/8/1 MSA-FFN blocks, respectively. *Partial* means that the last two stages, which contain most of the attention blocks, still use full attention. Vision Longformer does not have *Partial* version because its window size is set as 15 (comparable with the ViT(DeiT)/16 feature map size 14), and its attention mechanism in the last two stages is equivalent to full attention.}
\label{tab:attention_overall_timespace}
\end{table*}






%% file: app_vil_implement.tex
\section{Implementations and Efficiency of Vision Longformer In Practice}
\label{app:vil_implementation}
\begin{figure}[t!]
\includegraphics[width=0.48\textwidth]{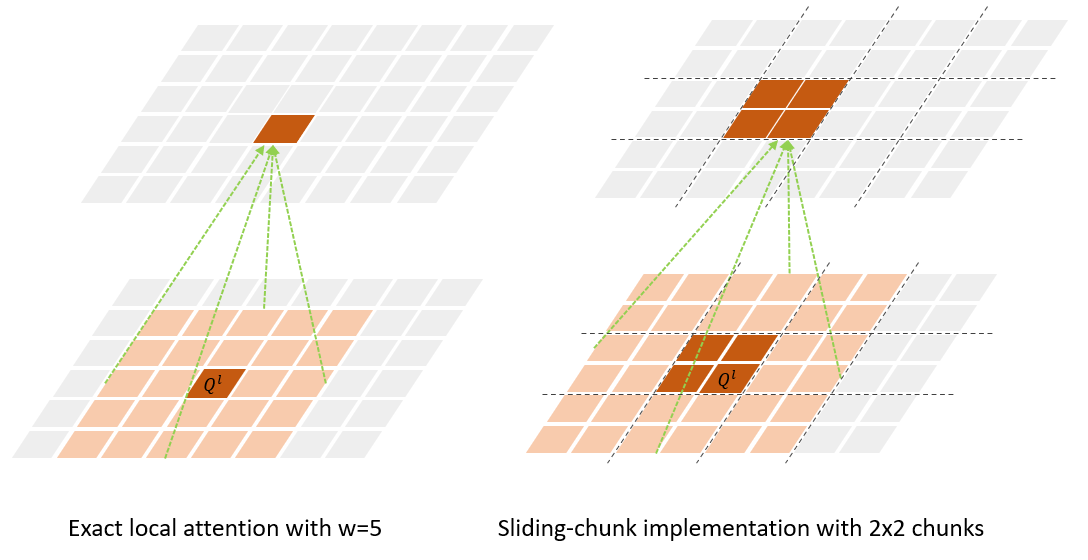}
\caption{The sliding-chunk implementation of Vision Longformer. This implementation (Right) lets one token attends to more tokens than the exact conv-like local attention (Left). Our sliding-chunk implementation has the choice to be exactly the same with the conv-like local attention (Left), by masking out tokens that should not be attended to. For chunks on the boundaries, our implementation supports both no padding and cyclic padding.}
\label{fig:vil_spacetime}
\vspace{-2mm}
\end{figure}
\begin{figure}[t!]
\includegraphics[width=0.48\textwidth]{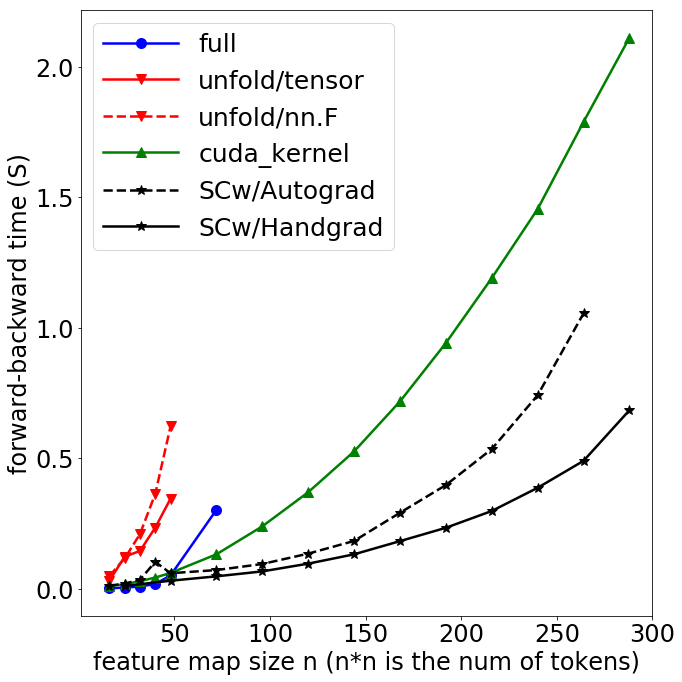}
\includegraphics[width=0.48\textwidth]{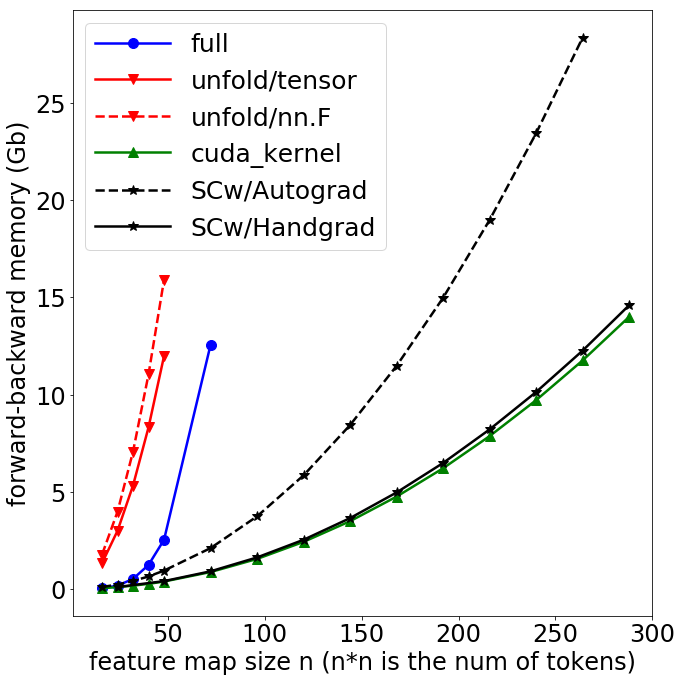}
\caption{Compare of running time (including forward and backward) and memory usage of different implementations of the conv-like attention in Vision Longformer. All of these implementations shown in the figures are mathematically equivalent, doing the exact conv-like sliding window attention with window size 17.}
\label{fig:vil_imps_compare}
\vspace{-2mm}
\end{figure}

There is a trivial implementation of the conv-like sliding window attention, in which we compute the full quadratic attention and then mask out non-neighbor tokens. This approach suffers from the quadratic complexity w.r.t. number of tokens (quartic w.r.t. feature map size), and is impractical for real use, as shown by the blue curve in Figure~\ref{fig:vil_imps_compare}. We only use it to verify the correctness of our other implementations. 

We have implemented Vision Longformer in three ways: 
\begin{enumerate}
    \item Using Pytorch's unfold function. We have two sub-versions: one using nn.functional.unfold (denoted as ``unfold/nn.F'') and the other using tensor.unfold (denoted as ``unfold/tensor''). As shown in Figure~\ref{fig:vil_imps_compare}, the ``unfold/tensor'' version (red solid line) is more efficient both in time and memory than the ``unfold/nn.F'' version (red dotted line). However, both of them are even slower and use more memory than the full attention!
    \item Using a customized CUDA kernel, denoted as ``cuda\_kernel''. We make use of the TVM, like what has done in Longformer~\cite{beltagy2020longformer}, to write a customized CUDA kernel for Vision Longformer. As shown in Figure~\ref{fig:vil_imps_compare}, the ``cuda\_kernel'' (green line) achieves the theoretical optimal memory usage. Its time complexity is also reduced to linear w.r.t. number of tokens (quadratic w.r.t. feature map size). However, since it's not making use of the highly optimized matrix multiplication libraries in CUDA, it's speed is still slow in practice.
    \item Using a sliding chunk approach, illustrated in Figure~\ref{fig:vil_spacetime}. For this sliding chunk approach, we have two subversions: one using Pytorch's autograd to compute backward step (denoted as ``SCw/Autograd") and the other writing a customized torch.autograd.Function with hand-written backward function (denoted as ``SCw/Handgrad"). Both sub versions of this sliding chunk approach are fully implemented with Pytorch functions and thus make use of highly optimized matrix multiplication libraries in CUDA. As shown in Figure~\ref{fig:vil_imps_compare}, both of them are faster than the ``cuda\_kernel'' implementation.
\end{enumerate}

In the sliding chunk approach, to achieve a conv-like local attention mechanism with window size $2 w + 1$, we split the feature map into chunks with size $w\times w$. Each chunk only attends to itself and its 8 neighbor chunks. The Pytorch Autograd will save 9 copies of the feature map (9 nodes in the computing graph) for automatic back-propagation, which is not time/memory efficient. The ``SCw/Handgrad" version defines a customized torch.autograd.Function with hand-written backward function, which greatly saves the memory usage and also speeds up the algorithm, as shown in Figure~\ref{fig:vil_imps_compare}. We would like to point out that the memory usage of the ``SCw/Handgrad" version is nearly optimal (very close to that of the ``cuda\_kernel''). Similar speed-memory trade-off with different implementations of local attention mechanism has been observed in the 1-D Longformer~\cite{beltagy2020longformer}, too; see Figure~1 in \cite{beltagy2020longformer}. We would like to point out that Image Transformer~\cite{parmar2018image} has an implementation of of 2-D conv-like local attention mechanism, which is similar to our ``SCw/Autograd" version. The Image Transformer~\cite{parmar2018image} applies it to the image generation task.

\begin{figure}[t!]
\includegraphics[width=0.48\textwidth]{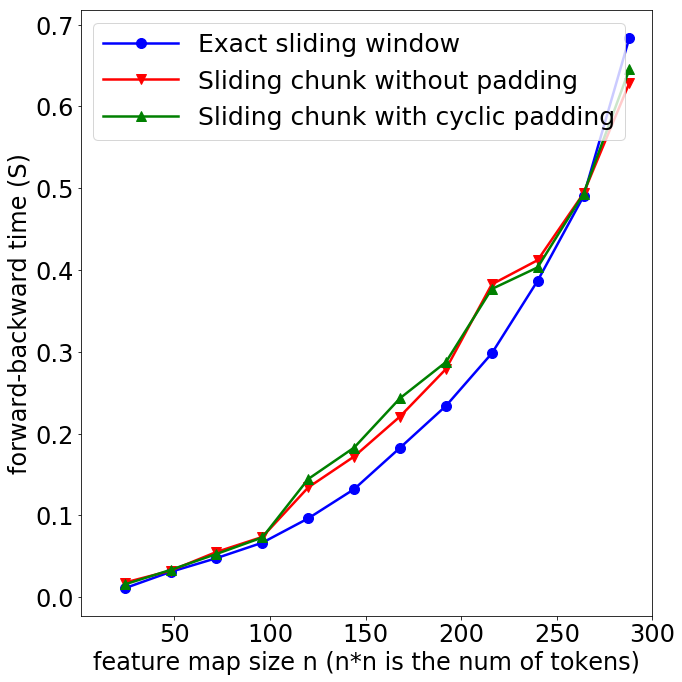}
\includegraphics[width=0.48\textwidth]{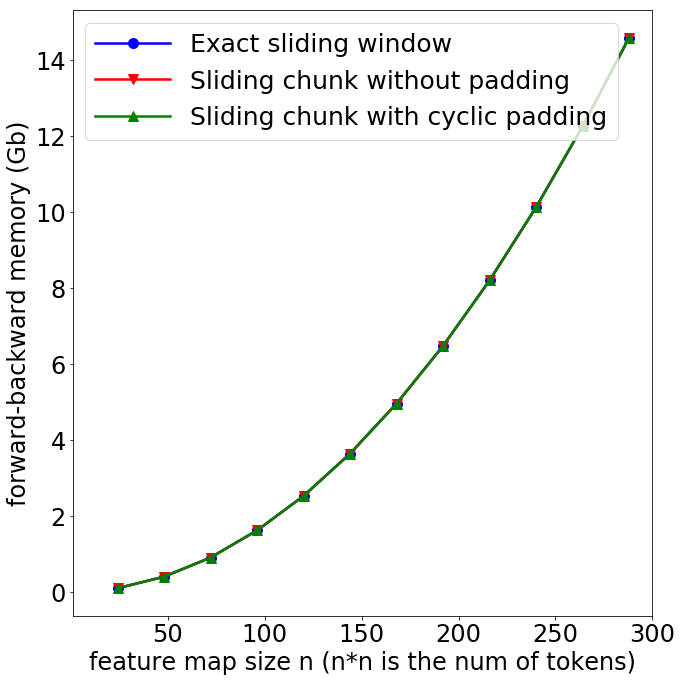}
\caption{Compare of three masking methods of our ``SCw/Handgrad'' implementation of conv-like local attention: exact conv-like sliding window attention, sliding chunk attention without padding for boundary chunks, and sliding chunk attention with cyclic padding for boundary chunks. The are nearly the same in terms of running time (including forward and backward) and memory usage. The window size is 17 and thus chunk size is 8.}
\label{fig:vil_padding_compare}
\vspace{-2mm}
\end{figure}
\begin{figure}[t!]
\includegraphics[width=0.48\textwidth]{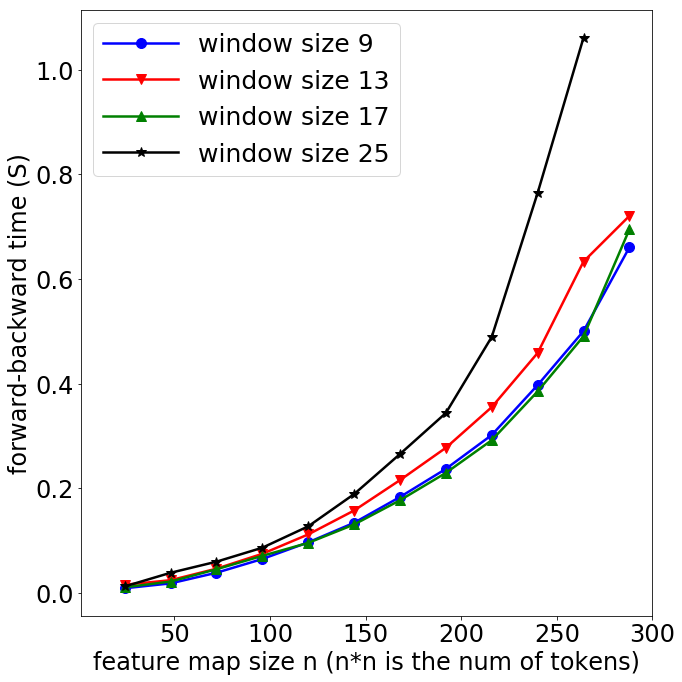}
\includegraphics[width=0.48\textwidth]{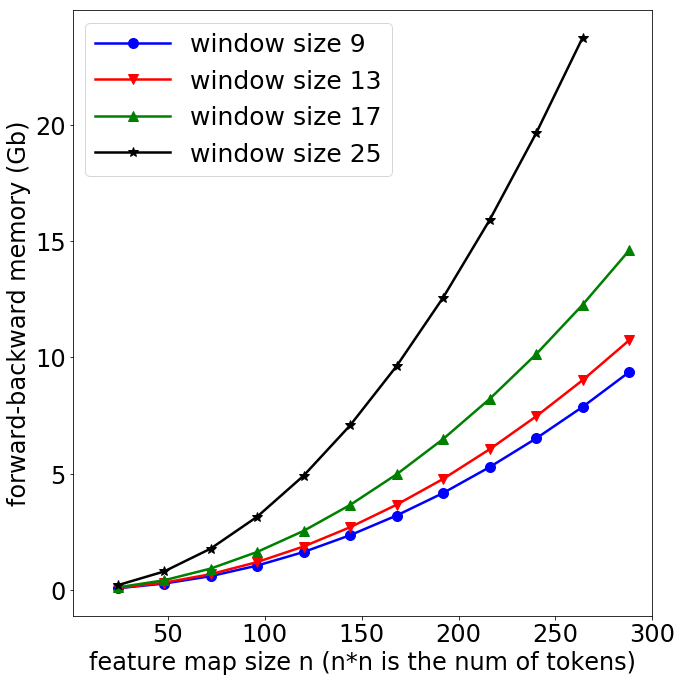}
\caption{Running time (including forward and backward) and memory usage of our ``SCw/Handgrad'' implementation of conv-like local attention (sliding chunk attention without padding mode) with different window sizes. The speed is not sensitive to the window size for small window sizes ($\le 17$) and the memory usage monotonically increases. }
\label{fig:vil_winsize_compare}
\vspace{-2mm}
\end{figure}

This sliding-chunk implementation (Figure~\ref{fig:vil_spacetime} Right) lets one token attends to more tokens than the exact conv-like local attention (Figure~\ref{fig:vil_spacetime} Left). Our sliding-chunk implementation has the choice to be 
\begin{enumerate}
    \item exactly the same with the conv-like local attention (Left), by masking out tokens that should not be attended to,
    \item sliding chunk without padding, in which the chunks on the boundary have less chunks to attend to,
    \item sliding chunk with cyclic padding, in which the chunks on the boundary still attend to 9 chunks with cyclic padded chunks. 
\end{enumerate}

\begin{figure}[t!]
\includegraphics[width=0.45\textwidth]{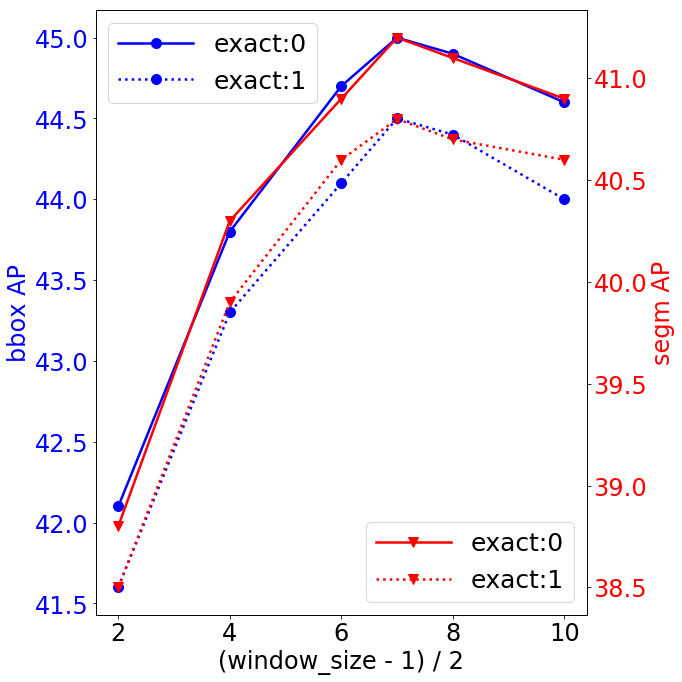}
\caption{``Sliding chunk without padding" performs consistently better than ``exact sliding window" for object detection with Mask R-CNN. All use the same ImageNet1K pre-trained checkpoint (\vil-Small-RPB in Table~\ref{tab:overall_comp_cls}).}
\label{fig:vil_exact}
\vspace{-2mm}
\end{figure}

Since these three masking methods only differ by the attention masks to mask out invalid tokens, their speed and memory usage are nearly the same, as shown in Figure~\ref{fig:vil_padding_compare}. For ImageNet
classification, we observe no obvious difference in top1 accuracy between ``exact sliding window" and ``sliding chunk without padding", while ``sliding chunk with cyclic padding" performs slightly worse most of the time. For object detection, we observe that ``sliding chunk without padding" performs consistently better than ``exact sliding window", as shown in Figure~\ref{fig:vil_exact}. Therefore, we make ``sliding chunk without padding'' as the default making method for Vision Longformer, although it sacrifices some translational invariance compared with ``exact sliding window". 

In Figure~\ref{fig:vil_winsize_compare}, we show the running time (including forward and backward) and memory usage of our ``SCw/Handgrad'' implementation of conv-like local attention (sliding chunk attention without padding mode) with different window sizes. We can see that the speed is not sensitive to the window size for small window sizes ($\le 17$) and the memory usage monotonically increases.

Finally, both the ``unfold/nn.F'' and the ``cuda\_kernel'' implementations support dilated conv-like attention. The customized CUDA kernel is even more flexible to support different dilations for different heads. The sliding-chunk implementation does not support this dilated conv-like attention. In this paper, we always use the sliding-chunk implementation due to its superior speed and nearly optimal memory complexity. 

In Figure~\ref{fig:vil_imps_compare}, \ref{fig:vil_padding_compare} and \ref{fig:vil_winsize_compare}, the evaluation is performed on a single multi-head self-attention module (MSA) with the conv-like local attention mechanism, instead of on the full multi-scale Vision Longformer. With this evaluation, we can clearly see the difference among different implementations of the conv-like local attention mechanism.

%% file: app_random_shift.tex
\section{Random-shifting strategy to improve training efficiency}
\label{app:random_swin_app}
\begin{figure}[t!]
\includegraphics[width=0.45\textwidth]{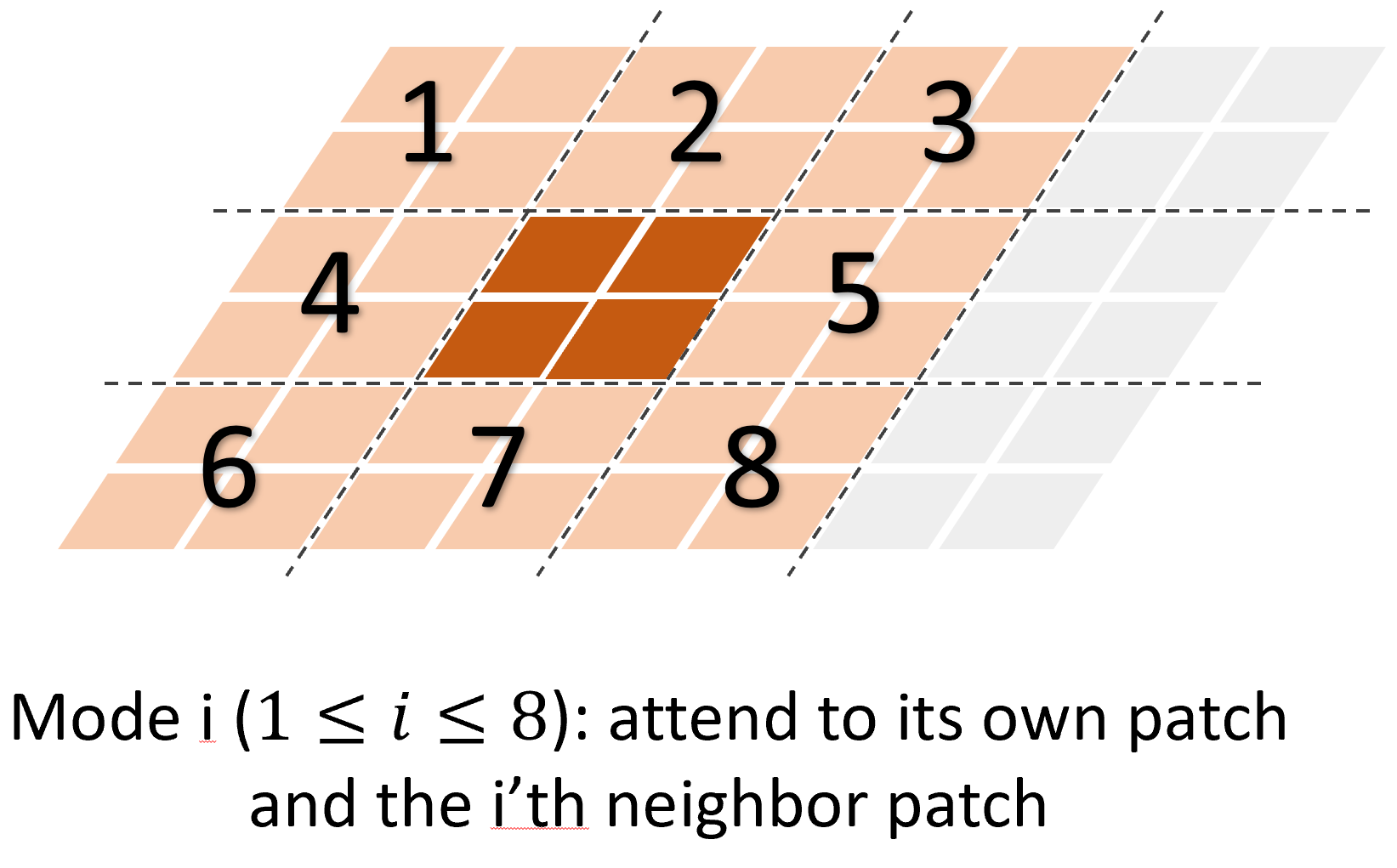}
\caption{Illustration of the 8 modes in the random-shifting training strategy. For mode $i$ ($1 <= i <=8$), the query chunk (dark brown) attends to itself and the $i$'th neighbor chunk.}
\label{fig:random_shift_mode}
\vspace{-2mm}
\end{figure}

\begin{figure}[t!]
\includegraphics[width=0.45\textwidth]{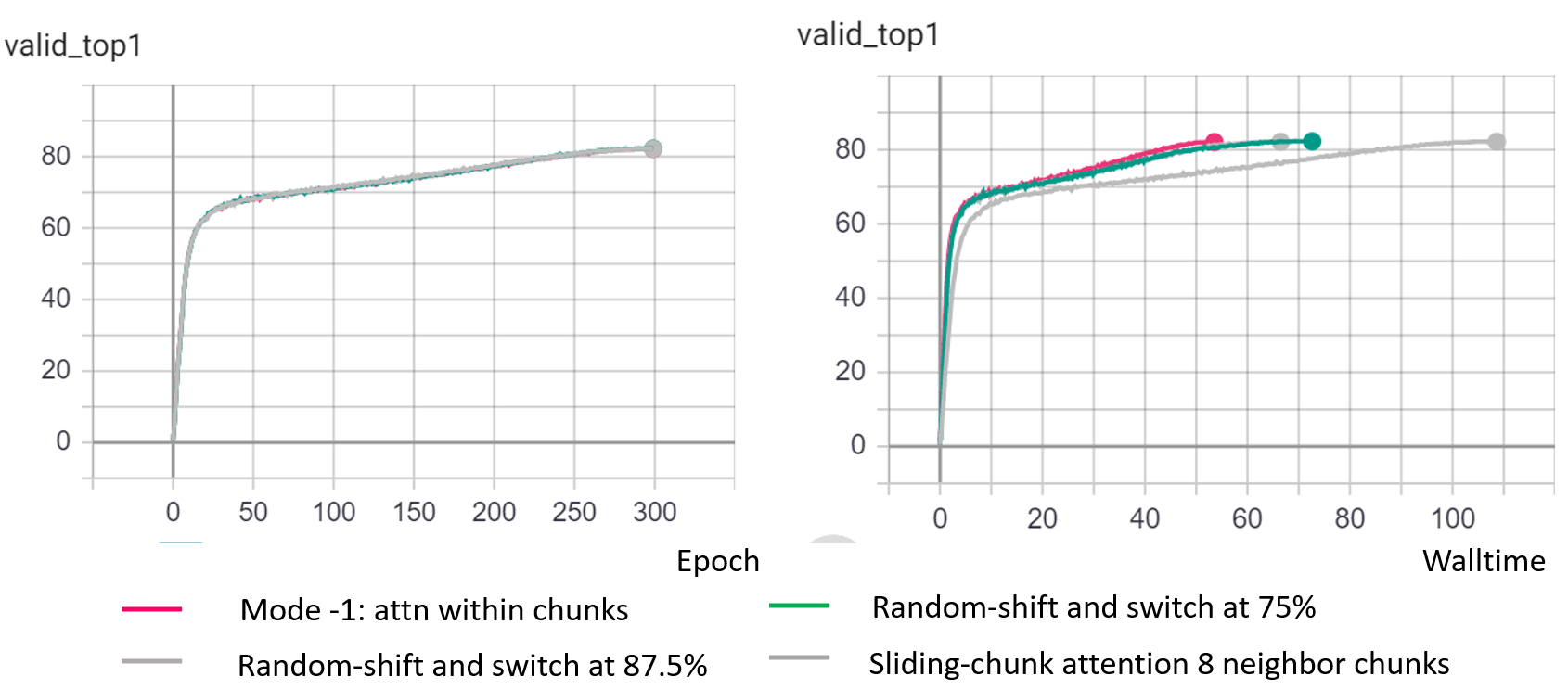}
\caption{The random-shifting strategy does not harm the model performance (Left), an accelerates the Vision Longformer training significantly (Right). When zooming in, the performance of ``random-shift and switch at 75\%" is slightly better than the ``Sliding-chunk attention with 8 neighbor chunks".}
\label{fig:random_shift_top1}
\vspace{-2mm}
\end{figure}

We propose the random-shifting training strategy for Vision Longformer, to further accelerate the training speed of Vision Longformer. More specifically, instead of attending to all 8 neighbor patches, one patch can only attend to itself and one random neighbor patch during training. To achieve this, we define 10 modes of the sliding-chunk local attention:
\begin{itemize}
    \item 0 (default): attend to itself and all 8 neighbor chunks,
    \item -1 : only attend to itself chunk,
    \item i ($1 <= i <=8$) : attend to itself chunk and the i'th neighbor chunk.
\end{itemize}
The ordering of the 8 neighbor patches is visualized in Figure~\ref{fig:random_shift_mode}. During training, we can randomly sample one mode from 1 to 8 and perform the corresponding random-shifting attention. We switch from the random-shifting mode to the default 8-neighbor mode after $x$\% training iterations, and this switch time $x$\% is a hyper-parameter with default value 75\%. This switch, can be seen as fine-tuning, is necessary to mitigate the difference of model's behavior during training and inference. As shown in Figure~\ref{fig:random_shift_top1}, this random-shifting training strategy accelerates the Vision Longformer training significantly, while not harming the final model performance. 

%% file: app_otherattns.tex
\section{Other Efficient Attention Mechanisms utilized in this work}
\label{app:efficientattn}
In this paper, we compare Vision Longformer with the following alternative choices of efficient attention methods.

\noindent\textbf{Pure global memory ($\textcolor{red}{a} = \text{global}$).} In Vision Longformer, see Figure~\ref{fig:vil_block} (Left), if we remove the local-to-local attention, then we obtain the pure global memory attention mechanism (called Global Attention hereafter). Its memory complexity is $\mathcal{O}(n_g(n_g + n_l))$, which is also linear w.r.t. $n_l$. However, for this pure global memory attention, $n_g$ has to be much larger than 1. In practice, we set different numbers of global tokens for different stages, as shown in Table~\ref{tab:attention_param_full}, with more global tokens in the first 2 stages and less in the last 2 stages. This setting makes the memory/computation complexity comparable with other attention mechanisms under the same model size.

\noindent\textbf{Linformer\cite{wang2020linformer} ($\textcolor{red}{a} = \text{LIN}$)} projects the $n_l \times d$ dimensional keys and values to $K \times d$ dimensions using additional projection layers, where $K \ll n_l$. Then the $n_l$ queries only attend to these projected $K$ key-value pairs. The memory complexity of Linformer is $\mathcal{O}(K n_l)$. We gradually increase $K$ (by 2 each time) and its performance gets nearly saturated at 256. Therefore, $K = 256$ is our choice for this Linformer attention, which turns out to be the same with the recommended value. Notice that Linformer's projection layer (of dimension $K \times n_l$) is specific to the current $n_l$, and cannot be transferred to higher-resolution tasks that have a different $n_l$. It is possible to transfer Linformer's weight by resizing feature maps of a different size to the original feature map size that Linformer is trained with and then applying the Linformer's projection. We do not explore this choice in this work. 

\noindent\textbf{Spatial Reduction Attention (SRA)~\cite{wang2021pyramid} ($\textcolor{red}{a} = \text{SRA}$)} is similar to Linformer, but uses a convolution layer with kernel size $R$ and stride $R$ to project the key-value pairs, hence resulting in $n_l / R^2$ compressed key-value pairs. Therefore, The memory complexity of SRA is $\mathcal{O}(n_l^2/R^2)$, which is still quadratic w.r.t. $n_l$ but with a much smaller constant $1/R^2$. When transferring the ImageNet-pretrained SRA-models to high-resolution tasks, SRA still suffers from the quartic computation/memory blow-up w.r.t. the feature map resolution. Pyramid Vision Transformer~\cite{wang2021pyramid} uses this SRA to build multi-scale vision transformer backbones, with different spatial reduction ratios ($R_1=8, R_2=4,R_3=2, R_4=1$) for each stage. With this PVT's setting, the key and value feature maps at all stages are essentially with resolution $\frac{H}{32} \times \frac{W}{32}$. This choice is able to scale up to image resolution $600 \times 1000$, but the memory usage is much larger than ResNet counterparts for $800 \times 1333$. 

In this paper, we benchmarked the performance of SRA/32 with SR ratios $R_1=8, R_2=4,R_3=2, R_4=1$ (same as PVT~\cite{wang2021pyramid}) and SRA/64 with SR ratios $R_1=16, R_2=8,R_3=4, R_4=2$ (two times more downsizing from that in PVT~\cite{wang2021pyramid}), as shown in Table~\ref{tab:attention_param_full}. The SRA/64 setting makes the memory usage comparable with other efficient attention mechanisms under the same model size, but introduces more parameters due to doubling the kernel size of the convolutional projection layer. 

\noindent\textbf{Performer~\cite{choromanski2020rethinking} ($\textcolor{red}{a} = \text{performer}$)} uses random kernel approximations to approximate the Softmax computation in MSA, and achieves a linear computation/memory complexity with respect to $n_l$ and the number of random features $K$. We use the default $K = 256$ orthogonal random features (OR) for Performer. The memory/space complexity of performer is  $\mathcal{O}(K d + n_l d + K n_l)$ while its computation/time complexity is $\mathcal{O}(K n_l d)$. For the time complexity, we ignore the complexity of generating the orthogonal random features, which in practice cannot be ignored during training. We refer to Section B.3 in \cite{choromanski2020rethinking} for a detailed discussion of theoretical computation/memory complexity of Performer.

One important technique in training Performer is to redraw the random features during training. In our ImageNet classification training, we adopt a heuristic adaptive redrawing schedule: redraw every $1 + 5 T$ iterations in Epoch $T$ ($T = 0, 1, ..., 299$). In our COCO object detection/segmentation training, the Performer is initialized from ImageNet pretrained checkpoint and thus there is no need to redraw very frequently in the initial training stage.Therefore, we redraw the random features every 1000 iterations in COCO object detection/segmentation training. 

\begin{table}[ht]
\begin{center}
\resizebox{\linewidth}{!}{
\begin{tabular}{l@{\hspace{1.5pt}}|c@{\hspace{3pt}}|c@{\hspace{3pt}}|c@{\hspace{3pt}}|c@{\hspace{3pt}}}
\toprule
Model & Stage1 & Stage2 & Stage3 & Stage4 \\
\midrule
 & \multicolumn{4}{c}{Window size $w$} \\
\hline
\vil-3stage & \multicolumn{2}{c|}{15} & 15 & 15 \\
\vil-4stage & 15 & 15 & 15 & 15 \\
\vil-4stage (384) & 13 & 17 & 25 & 25 \\
\midrule
 & \multicolumn{4}{c}{Number of global tokens $n_g$} \\
\hline
Global-3stage & \multicolumn{2}{c|}{256} & 64 & 16 \\
Global-4stage & 256 & 256 & 64 & 16 \\
\midrule
 & \multicolumn{4}{c}{Projection Dimension $K$} \\
\hline
Linformer-3stage & \multicolumn{2}{c|}{256} & 256 & 256 \\
Linformer-4stage & 256 & 256 & 256 & 256 \\
\midrule
 & \multicolumn{4}{c}{Spatial reduction ratio $R$} \\
 \hline
SRA/64-3stage & \multicolumn{2}{c|}{8} & 4 & 2 \\
SRA/64-4stage & 16 & 8 & 4 & 2 \\
SRA/32-3stage & \multicolumn{2}{c|}{4} & 2 & 1 \\
SRA/32-4stage & 8 & 4 & 2 & 1 \\
\midrule
 & \multicolumn{4}{c}{Number of orthogonal random features $K$} \\
\hline
Performer-3stage & \multicolumn{2}{c|}{256} & 256 & 256 \\
Performer-4stage & 256 & 256 & 256 & 256 \\
\bottomrule
\end{tabular}
}
\end{center}
\caption{Attention mechanism specific hyper-parameters. See Appendix~\ref{app:efficientattn} for the details of these attention mechanisms.}
\label{tab:attention_param_full}
\end{table}